\documentclass{article}
\usepackage{ijcai26}

\usepackage{times}
\usepackage{soul}
\usepackage{url}
\usepackage[hidelinks]{hyperref}
\usepackage[utf8]{inputenc}
\usepackage[small]{caption}
\usepackage{graphicx}
\usepackage{amsmath}
\usepackage{amsthm}
\usepackage{amssymb}
\usepackage{booktabs}
\usepackage{algorithm}
\usepackage{algorithmic}
\usepackage[switch]{lineno}

\usepackage[hidelinks]{hyperref}
\usepackage{booktabs}
\usepackage{mathtools}
\usepackage{multirow}
\usepackage[table]{xcolor}
\usepackage{placeins}
\usepackage{amssymb}
\usepackage{caption}
\usepackage{adjustbox}
\usepackage{xcolor,colortbl}
\usepackage{pifont}
\usepackage[most]{tcolorbox}
\newcommand{\cmark}{\ding{51}}  
\newcommand{\xmark}{\ding{55}}  
\newtcolorbox{promptbox}{
    colback=gray!10,
    colframe=gray!40,
    boxrule=0.5pt,
    arc=6pt,
    left=6pt,
    right=6pt,
    top=6pt,
    bottom=6pt,
    fontupper=\small
}
\newtcolorbox{outputbox}{
    colback=blue!6,
    colframe=blue!40,
    boxrule=0.5pt,
    arc=6pt,
    left=6pt,
    right=6pt,
    top=6pt,
    bottom=6pt,
    fontupper=\small
}
\newtcolorbox{logicbox}{
    colback=green!6,
    colframe=green!40,
    boxrule=0.5pt,
    arc=6pt,
    left=6pt,
    right=6pt,
    top=6pt,
    bottom=6pt,
    fontupper=\small
}

\usepackage{listings}
\title{Neurosymbolic Framework for Concept-Driven Logical Reasoning in Skeleton-Based Human Action Recognition}

\let\oldtwocolumn\twocolumn
\renewcommand\twocolumn[1][]{%
    \oldtwocolumn[{#1}{
    \begin{center}
           \includegraphics[width=0.8\textwidth]{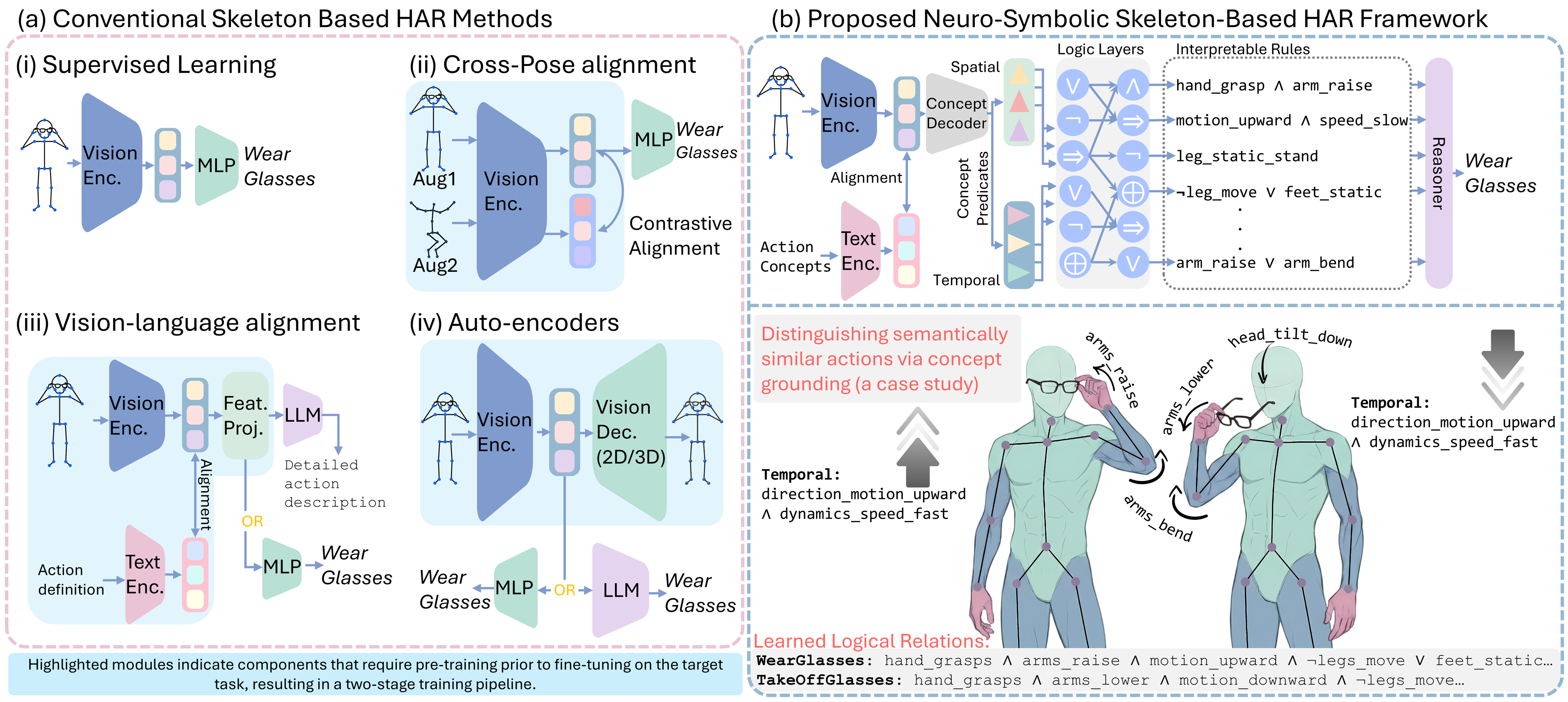}
           \captionof{figure}{\textbf{Our proposed concept-grounded neurosymbolic reasoning for skeleton-based activity recognition}: Grounding spatial–temporal motion concepts in differentiable logic to learn interpretable and compositional action rules.
           }
           \label{fig:fig1}
        \end{center}
    }]
}

\author{
Talha Ilyas$^{1,2}$
\and
Deval Mehta$^{2,3\ast}$\and
Zongyuan Ge$^{2,3}$
\affiliations
$^1$Department of ECSE, Faculty of Engineering, Monash University, Australia\\
$^2$AIM for Health Lab, Faculty of Information Technology, Monash University, Australia\\
$^3$Department of DSAI, Faculty of Information Technology, Monash University, Australia
\emails
\{talha.ilyas, deval.mehta\}@monash.edu
}

\begin{document}

\maketitle
\renewcommand{\thefootnote}{\fnsymbol{footnote}}
\footnotetext[1]{Corresponding author}

\begin{abstract}
    Skeleton-based human activity recognition has achieved strong empirical performance, yet most existing models remain black boxes and difficult to interpret. In this work, we introduce a neurosymbolic formulation of skeleton-based HAR that reframes action recognition as concept-driven first-order logical reasoning over motion primitives. Our framework bridges representation learning and symbolic inference by grounding first-order logic predicates in learnable spatial and temporal motion concepts. Specifically, we employ a standard spatio-temporal skeleton encoder to extract latent motion representations, which are then mapped to interpretable concept predicates via a spatio-temporal concept decoder that explicitly separates pose-centric and dynamics-centric abstractions. These concept predicates are composed through differentiable first-order logic layers, enabling the model to learn human-readable logical rules that govern action semantics. To impose semantic structure on the learned concepts, we align skeleton representations with LLM-derived descriptions of atomic motion primitives, establishing a shared conceptual space for perception and reasoning. Extensive experiments on NTU RGB+D 60/120 and NW-UCLA demonstrate that our approach achieves competitive recognition performance while providing explicit, interpretable explanations grounded in logical structure. Our results highlight neurosymbolic reasoning as an effective paradigm for interpretable spatio-temporal action understanding. \href{https://github.com/Mr-TalhaIlyas/REASON}{Code}$^{\dagger}$.\footnotetext[2]{Extended version of paper accepted at IJCAI 2026}
\end{abstract}
    
\section{Introduction}

Skeleton-based human activity recognition (HAR) provides an efficient, privacy-preserving alternative to video-based methods by focusing on joint coordinates that isolate human structure from environmental noise~\cite{duan2022revisiting,liu2023skeleton,asif2023deepactsnet}. While deep learning has advanced classification accuracy, these models remain black boxes, problematic for high-stakes domains such as healthcare and rehabilitation where transparent decision-making is essential~\cite{ilyas2025privacy,mehta2023privacy}.

Neurosymbolic AI has gained traction for combining neural perception with formal logic interpretability~\cite{garcez2002neural,bhuyan2024neuro}. However, its application to skeleton-based HAR remains largely unexplored due to fundamental integration challenges. This work addresses these challenges to achieve interpretable skeleton-based HAR.

\noindent\textbf{Symbol Grounding.} A key challenge is bridging continuous joint trajectories and discrete symbolic predicates. Traditional joint heuristics (e.g., knee angle $< 90^o$ for \textit{bent knee}) \cite{okamoto2024hierarchical} are brittle, as small pose errors or variations in viewpoint and execution can prevent predicate activation for reliable HAR. Inspired by Concept Bottleneck Models (CBMs)~\cite{koh2020concept}, we instead learn abstract motion concepts and construct an LLM-generated concept bank of spatial primitives (e.g., \texttt{arm\_raise}, \texttt{knee\_bend}) that serve as symbolic predicates. We also enforce semantic grounding by aligning GCN-extracted skeleton features with pretrained text encoder based concept embeddings.

\noindent\textbf{Temporal Dynamics.} Actions are inherently temporal, varying in speed, duration, and execution order. \textsc{WearingGlasses} and \textsc{TakeOffGlasses} share similar static poses (\texttt{hands near the face}, \texttt{arm engagement}) but differ in temporal dynamics: the former involves grasping and upward motion, while the latter requires release and downward motion. Purely spatial logic cannot capture such distinctions, and incorporating temporal logical rules into neurosymbolic frameworks is challenging. To address this, we introduce explicit temporal concepts capturing motion direction (e.g., \texttt{motion\_upward}, \texttt{motion\_downward}), hand state transitions (e.g., \texttt{hand\_grasp}, \texttt{hand\_release}), and execution order (e.g., \texttt{seq\_hands\_first}). Our Spatio-Temporal Concept Decoder (STC-Decoder) then independently decodes spatial and temporal concepts, enabling distingushing semantically similar actions such as $(\texttt{arm\_raise} \land \texttt{hand\_grasp} \land \texttt{motion\_upward}) \Rightarrow \textsc{WearGlasses}$ versus $(\texttt{arm\_lower} \land \texttt{hand\_release} \land \texttt{motion\_downward}) \Rightarrow \textsc{TakeOffGlasses}$ (Fig~\ref{fig:fig1}(b)).

\noindent\textbf{End-to-End Learning.} Discrete Boolean operators block gradient propagation, preventing joint optimization of perception and reasoning components~\cite{schrouff2021best}. While differentiable fuzzy logic \cite{barbiero2023interpretable} shows promise, scalability to large-scale, fine-grained skeleton-based HAR remains open. We employ differentiable logic layers with soft AND/OR operations \cite{perfilieva2002logical} and gradient grafting \cite{wang2023learning}, enabling discrete inference with end-to-end differentiability. Final predictions aggregate triggered rules via learned weights, defining actions as logical compositions of motion concepts.

\vspace{0.3em}
Together, these components form an end-to-end neurosymbolic framework (Fig~\ref{fig:fig1}(b)) in which a GCN encoder extracts spatio-temporal features with sequence–concept alignment, the STC-Decoder predicts interpretable concepts, and differentiable logic layers compose them via first-order operations ($\land$, $\lor$, $\neg$) to derive action predictions for action recognition. Grounding predictions in logical rules over concepts provides both local (concepts) and global interpretability (rules). Experiments on NTU RGB+D 60/120~\cite{shahroudy2016ntu,liu2019ntu} and NW-UCLA~\cite{wang2014cross} demonstrate we achieve competitive performance with transparent explanations. To summarize, our \textbf{major contributions} are:
\begin{itemize}
     \item \textsc{REASON} (\textbf{R}ule-based \textbf{E}xplainable \textbf{A}ction via \textbf{S}ymbolic c\textbf{ON}cepts), a neurosymbolic framework for skeleton-based HAR that formulates action recognition as logical reasoning over learnable motion concepts and integrates concept-grounded representations with differentiable first-order logic for end-to-end training.
    \item A Spatio-Temporal Concept Decoder (STC-Decoder) that extracts spatial body-configuration and temporal motion-dynamics concepts, yielding interpretable abstractions robust to execution variability.
    \item A concept bank of spatial and dynamic concepts for actions available in NTU RGB+D and NW-UCLA.
\end{itemize}


\section{Related Work}

\noindent\textbf{Skeleton-based HAR.}
ST-GCN~\cite{yan2018spatial} established the GCN paradigm by modeling joints as graph nodes with spatio-temporal convolutions. Subsequent work (Fig~\ref{fig:fig1}(a)-(i),(ii)) addresses topology rigidity: 2s-AGCN~\cite{shi2019two} learns adaptive joint relationships, CTR-GCN~\cite{chen2021channel} refines topology channel-wise, and InfoGCN~\cite{chi2022infogcn} incorporates information-theoretic objectives. Recent architectures like HD-GCN~\cite{lee2023hierarchically} and BlockGCN~\cite{zhou2024blockgcn} capture hierarchical interactions. Self-supervised approaches (Fig~\ref{fig:fig1}(a)-(iv)) including SkeletonMAE~\cite{wu2023skeletonmae} and MotionBERT~\cite{zhu2023motionbert} improve generalization via masked reconstruction. Despite strong performance, these models lack interpretable action semantics.

\noindent\textbf{Vision-Language for HAR.}
Recent works have used LLMs for skeleton-based HAR (Fig~\ref{fig:fig1}(a)-(iii),(iv)). SUGAR~\cite{ye2025sugar} and LLM-AR~\cite{qu2024llms} leverage LLM priors for classification or incorporate action prompts~\cite{xiang2023generative}. However, these methods yield generic, non-instance-specific explanations and cannot distinguish execution variants (e.g., arms-first vs. legs-first), and are impractical for edge deployment. In contrast, our framework learns efficient, compositional logical rules over motion predicates that capture multiple execution pathways.

\begin{figure*}[!ht]
    \centering
    \includegraphics[width=0.85\textwidth]{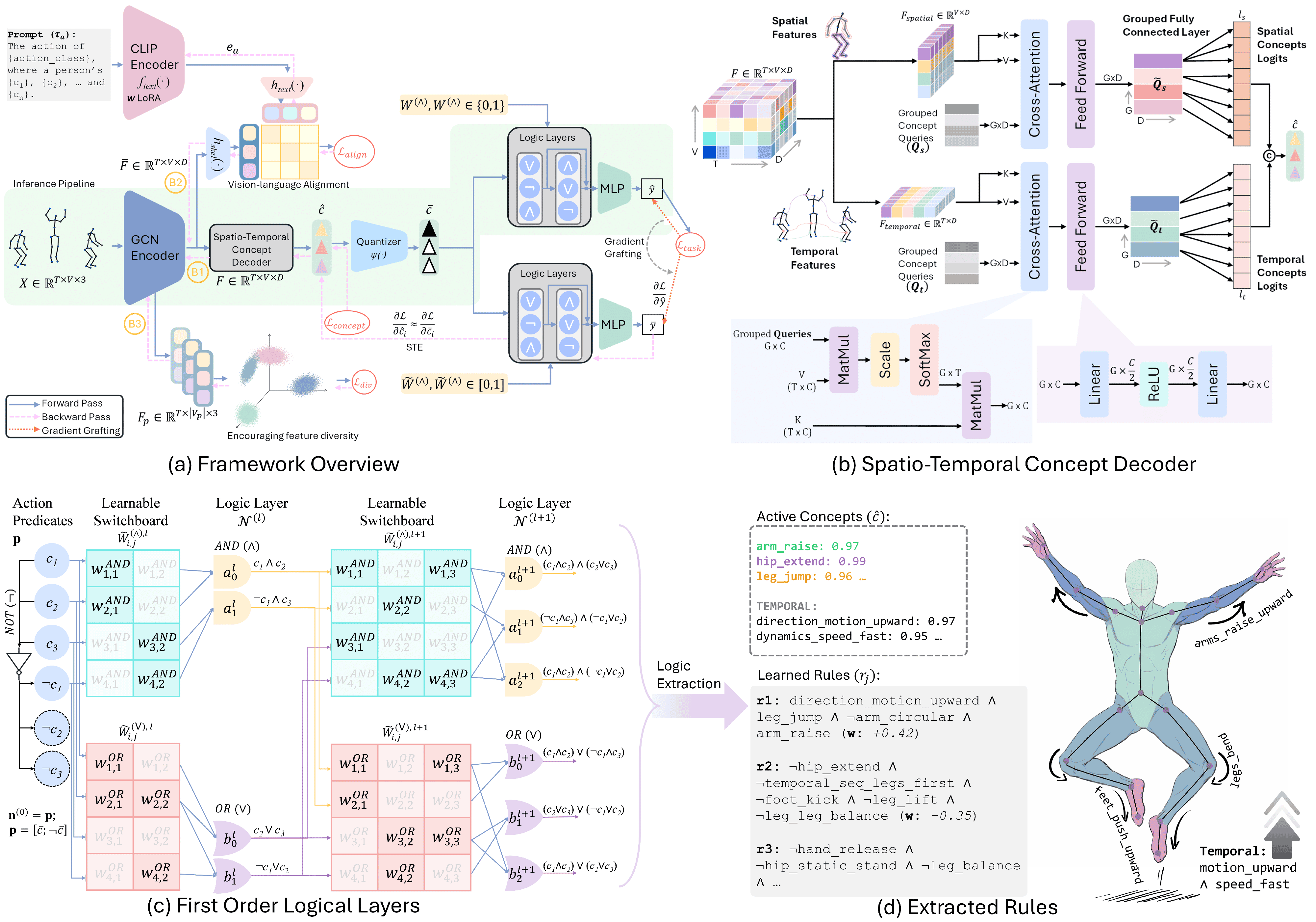}
    \caption{(a) Overall neurosymbolic, concept-grounded framework for skeleton-based human activity recognition. 
    (b) STC-Decoder for decoupling encoder features into spatial and temporal streams to learn spatio-temporal motion concepts. (c) First-order logic layers, for learning logical predicates and compositional rules over predicted concepts. (d) Example visualization for a \textsc{Jump} action, showing top activated concepts and learned rules.}
    \label{fig:model}
\end{figure*}

\noindent\textbf{Explainable HAR and Neurosymbolic AI.}
Prior interpretable HAR work has focused on RGB video via post-hoc concept attribution~\cite{kowal2024understanding} or neurosymbolic pipelines with expert-defined rules~\cite{okamoto2024hierarchical}, inheriting privacy concerns and sensitivity to environmental noise. While neurosymbolic systems enable traceable reasoning~\cite{hitzler2022neuro}, skeleton data exhibit substantial subject and viewpoint variability, making hand-crafted rules brittle. We address this with learnable concept predicates combined with differentiable logic layers to discover action-defining rules from data.

\noindent\textbf{Concept Bottleneck Models.}
CBMs constrain representations to interpretable concept layers, enabling intervention and debugging~\cite{koh2020concept}, with extensions like Logic CBMs~\cite{vemuri2025logiccbms} incorporating differentiable logic for concept compositions. Recent work further automates concept bank construction using LLM priors~\cite{mehta2025interpretable,jiang2025enhancing}, reducing manual effort. Despite success in static image classification tasks~\cite{jiang2025wise}, CBMs have not addressed skeleton-based HAR, which requires joint modeling of spatial and temporal dynamics. We bridge this gap by formulating skeleton-based HAR as logical reasoning over learnable spatio-temporal motion concepts.

\section{Method}
\label{sec:method}

Our work centres on the question: \textbf{Can concepts serve as an intermediate representation for neurosymbolic skeleton-based HAR?} We address this by first constructing a concept bank of spatial and temporal motion primitives (Sec.~\ref{sec:concept_bank}), and then developing a framework to predict concepts from skeleton sequences (Sec.~\ref{sec:skeleton_rep}\&~\ref{sec:stc_decoder}) and composing them via differentiable logic layers (Sec.~\ref{sec:logic_layers}\&~\ref{sec:action_classification}) for interpretable HAR.

\subsection{Concept Bank Generation}
\label{sec:concept_bank}

Unlike image classification where concepts correspond to static attributes (e.g., a zebra has \texttt{stripes}), action recognition requires capturing spatio-temporal dynamics. Furthermore, execution variability within an action (e.g., arm-assisted v/s arm-free \textsc{jump}) makes manual concept enumeration infeasible. We address this by construcing a concept bank $\mathcal{C} = \mathcal{C}_\text{s} \cup \mathcal{C}_\text{t} \cup \mathcal{C}_\text{int}$ of spatial, temporal, and interactive motion concepts through a three-stage LLM-based pipeline.

\subsubsection{Fine-grained Action Description Generation} 
Following HAKE~\cite{li2022hake}, we decompose actions into six body parts: $\mathcal{P}=\{\texttt{head},\,\texttt{hand},\,\texttt{arm},\,\texttt{hip},\,\texttt{leg},\,\texttt{foot}\}$. We then prompt an LLM (GPT-3) to generate part-wise motion descriptions $\mathcal{D}_a = \{d_a^{(p)}\}_{p \in \mathcal{P}}$ for each action $a$. For example, \textsc{Jump} yields: \{\texttt{Leg}: bends at knees then pushes explosively; \texttt{Arm}: extends overhead; \texttt{Foot}: pushes off ground\}.

\subsubsection{From Action Descriptions to Abstract Concepts} 
Generated raw descriptions are action-specific. Thus, using them directly could cause vocabulary explosion~\cite{jiang2025enhancing}. Hence, we convert descriptions into canonical concepts by extracting \{\texttt{verb-direction-modifier}\} patterns from all actions, then embedding them via a sentence encoder $\phi$, and clustering them with $k$-means. Cluster representatives form spatial concepts $\mathcal{C}_\text{s}$ (e.g., \texttt{arm\_raise}, \texttt{leg\_jump}, \texttt{hand\_grasp}). Additionally, we define temporal concepts $\mathcal{C}_\text{t}$ capturing motion direction, sequence, and dynamics (e.g., \texttt{motion\_upward}, \texttt{hands\_first}, \texttt{speed\_fast}), and an interaction concept for mutual actions.

\subsubsection{Concept-Action Association and Refinement} 
We construct an association matrix $\mathbf{M} \in \{0, 1\}^{|\mathcal{A}| \times |\mathcal{C}|}$ where $\mathbf{M}_{a,c} = 1$ indicates concept $c$ is active for action $a$. Using LLaMA-3.2 7B~\cite{grattafiori2024llama}, we match each action's descriptions to semantically appropriate concepts, allowing multiple concepts per body part for compositional movements (e.g., \textsc{Jump} activates both \texttt{leg\_squat} and \texttt{leg\_jump}). To ensure discriminability, we verify unique concept signatures and invoke refinement for ambiguous pairs (e.g., \textsc{WearGlasses} vs. \textsc{TakeOffGlasses} are distinguished via opposing direction concepts, see Figure~\ref{fig:fig1}). Our final concept bank comprises $|\mathcal{C}| = 74$ concepts (52 spatial + 21 temporal + 1 interaction). Implementation details, prompts, pseudocode and the complete vocabulary are provided in Appendix~\ref{app:concept_bank}.

\subsection{Skeleton Representation Learning}
\label{sec:skeleton_rep}

After constructing the concept bank, we describe the overall framework (Fig~\ref{fig:model}(a)). Skeleton sequences are first encoded using a GCN and aligned with concept semantics via a concept-grounded text encoder and a contrastive objective.

\textbf{Skeleton Encoder}
\label{sec:skeleton_encoder}
Let $\mathbf{X} \in \mathbb{R}^{T \times V \times 3}$ denote an input skeleton sequence with $T$ frames, $V$ joints and (\textit{x,y,z}) coordinates. We model the skeleton as a spatio-temporal graph and process it through a GCN backbone (e.g., Hyper-GCN~\cite{zhou2025adaptive}) with multi-scale temporal convolutions. Crucially, we retain full spatio-temporal resolution without pooling, outputting features $\mathbf{F} \in \mathbb{R}^{T \times V \times D}$ to preserve fine-grained motion information for concept prediction.

The encoder feeds into three branches (Fig~\ref{fig:model}(a)): \textbf{(B1)} the STC-Decoder for concept prediction, \textbf{(B2)} a text alignment branch where pooled features $\bar{\mathbf{F}} = \texttt{Pool}(\mathbf{F})$ are aligned with concept embeddings, and \textbf{(B3)} a part-divergence branch promoting discriminative representations across anatomical regions.

\textbf{Body-Part Feature Divergence}
\label{sec:feat_div}
We partition joints into subsets $\{\mathcal{V}_p\}_{p \in \mathcal{P}}$ and extract part-specific features $\mathbf{F}_p \in \mathbb{R}^{T \times |\mathcal{V}_p| \times D}$. To prevent feature collapse across body parts, we introduce a divergence loss that penalizes high cosine similarity between pooled part representations. Let $\bar{\mathbf{f}}_p = \texttt{Pool}(\mathbf{F}_p) \in \mathbb{R}^D$ denote the feature of part ($p$). The loss minimizes off-diagonal similarities across all part pairs:
\begin{equation}
\mathcal{L}_{\text{div}}
=
\frac{1}{|\mathcal{P}|(|\mathcal{P}|-1)}
\sum_{p \neq q}
\text{ReLU}
\!\left(
\frac{\bar{\mathbf{f}}_p^\top \bar{\mathbf{f}}_q}
{\|\bar{\mathbf{f}}_p\|\,\|\bar{\mathbf{f}}_q\|}
\right)
\end{equation}

This regularization encourages each body part to learn discriminative features aligned with its corresponding concept subset $\mathcal{C}_p \subset \mathcal{C}_\text{spatial}$.


\textbf{Concept-Grounded Text Encoder}
\label{sec:text_encoder}
To ground skeleton features in semantic space, we encode textual descriptions using CLIP~\cite{radford2021learning} with LoRA fine-tuning~\cite{hu2022lora}. Crucially, rather than generic action labels, we construct concept-grounded prompts incorporating motion primitives from our concept bank. For action $a$ with associated concepts $\{c_i\}$ from matrix $\mathbf{M}$, we construct:

\begin{quote}
\small
\textit{``A person performing [action], where the [part$_1$] [concept$_1$], the [part$_2$] [concept$_2$], ...''}
\end{quote}

\noindent For example, \textsc{Jump} yields: ``A person performing \textsc{jump}, where the \texttt{arms} \emph{swing upward}, the \texttt{knees} \emph{bend}, and the \texttt{body} \emph{moves upward} briefly.'' This formulation ensures text embeddings $\mathbf{e}_a = f_\text{text}(\tau_a)$ encode the same compositional semantics as the concept vocabulary.

We then align skeleton and text representations via contrastive learning. Projected skeleton features $\mathbf{z} = h_\text{skel}(\bar{\mathbf{F}})$ and concept embeddings $\mathbf{t} = h_\text{text}(\mathbf{e}_a)$ are aligned through:
\begin{equation}
    \mathcal{L}_\text{align} = -\frac{1}{B} \sum_{i=1}^{B} \log \frac{\exp(\mathbf{z}_i^\top \mathbf{t}_i / \tau)}{\sum_{j=1}^{B} \exp(\mathbf{z}_i^\top \mathbf{t}_j / \tau)}
\end{equation}
where $\tau$ is a learnable temperature. This objective encourages skeleton features to cluster with their concept-grounded descriptions, reinforcing the compositional structure.

\subsection{STC-Decoder}
\label{sec:stc_decoder}

The spatio-temporal concept decoder (STC-Decoder, Fig.~\ref{fig:model}b) maps continuous skeleton sequences to discrete concept activations, bridging skeleton representations and symbolic predicates. As actions activate multiple concepts, we adopt a multi-label design with separate spatial and temporal pathways.

\noindent\textbf{Spatial-Temporal Feature Decoupling.}
Given encoder features $\mathbf{F} \in \mathbb{R}^{T \times V \times D}$, we decouple them through dimension-wise aggregation:
\begin{equation}
    \mathbf{F}_\text{s} = \frac{1}{T} \sum_{t=1}^{T} \mathbf{F}[t, :, :] \in \mathbb{R}^{V \times D}, \,
    \mathbf{F}_\text{t} = \frac{1}{V} \sum_{v=1}^{V} \mathbf{F}[:, v, :] \in \mathbb{R}^{T \times D}
\end{equation}
Spatial features $\mathbf{F}_\text{s}$ preserve per-joint information by aggregating across time, while temporal features $\mathbf{F}_\text{t}$ retain motion dynamics by aggregating across joints.

\noindent\textbf{Group-Decoding with Concept Queries.}
Predicting all $|\mathcal{C}|$ concepts jointly poses an optimization challenge: concepts exhibit heterogeneous semantic relationships, some are mutually exclusive (\texttt{motion\_upward} vs. \texttt{motion\_downward}), others frequently co-occur (\texttt{arm\_raise}, \texttt{hand\_grasp}), and many are conditionally dependent on body-part context. Learning these complex inter-concept relationships in a single framework is difficult.

We address this by partitioning concepts into $G$ groups, where each group predicts $g = \lceil |\mathcal{C}| / G \rceil$ concepts. Rather than manually assigning semantic roles, we let each group \emph{learn} to specialize through end-to-end training, groups naturally discover coherent concept subsets (e.g., related body-part configurations or temporal patterns) that exhibit homogeneous co-occurrence, making each sub-problem easier to optimize.


We initialize separate learnable query matrices $\mathbf{Q}_\text{s} \in \mathbb{R}^{G_\text{s} \times D}$ and $\mathbf{Q}_\text{t} \in \mathbb{R}^{G_\text{t} \times D}$ for spatial and temporal branches. Each branch performs cross-attention followed by a feed-forward network, where group queries learn to selectively aggregate skeleton information relevant to their respective concept subsets (Appendix~\ref{app:stc_decoder}).

\noindent\textbf{Group Fully-Connected Layer.}
A Group-FC layer~\cite{ridnik2023ml} projects the refined group queries to per-concept predictions. Each group $k \in \{0, \ldots, G-1\}$ maintains its own projection $\mathbf{W}_k \in \mathbb{R}^{D \times g}$, responsible for predicting $g$ concepts. For concept $i$, we determine its group index $k = \lfloor i / g \rfloor$ and position within the group $j = i \mod g$, then compute:
\begin{equation}
    \ell_i = \tilde{\mathbf{Q}}[k, :] \cdot \mathbf{W}_k[:, j]
\end{equation}
This factorization encourages each group query to specialize in its concept subset through gradient-driven learning. Aggregating spatial and temporal logits yields:
\begin{equation}
    \hat{\mathbf{c}} = \sigma([\boldsymbol{\ell}_\text{s}; \boldsymbol{\ell}_\text{t}]) \in [0, 1]^{|\mathcal{C}|}
\end{equation}
where $\hat{c}_i$ denotes the predicted probability of concept $c_i$, providing soft symbolic grounding for the logic layers.

\noindent\textbf{Concept Prediction Loss.} Concept prediction is supervised using the action–concept matrix $\mathbf{M}$. For an action label $a$, the target concept vector is $\mathbf{c}^* = \mathbf{M}[a,:] \in \{0,1\}^{|\mathcal{C}|}$, and training minimizes binary cross-entropy $\mathcal{L}_{\text{concept}} = \mathrm{BCE}(\mathbf{c}^*, \hat{\mathbf{c}})$. This supervision grounds continuous skeleton features into interpretable concept activations $\hat{\mathbf{c}}$, which serve as symbolic predicates for logical composition.

\subsection{Differentiable First-Order Logic Reasoning}
\label{sec:logic_layers}

Symbolic reasoning requires binary predicates, whereas our predicted concepts $\hat{\mathbf{c}} \in [0,1]^{|\mathcal{C}|}$ provide soft activations over the motion vocabulary. We therefore binarize concept predictions and compose them using first-order logic, while also employing continuous relaxations to enable end-to-end learning and preserve discrete, interpretable rules.

\subsubsection{Concept Binarization}

Concept leakage~\cite{mahinpei2021promises} occurs when downstream layers exploit spurious information in soft activations rather than intended semantics. We therefore employ a concept activation binarizer which applies a threshold $\bar{c}_i = \mathbb{1}[\hat{c}_i > 0.5]$, yielding $\bar{\mathbf{c}} \in \{0,1\}^{|\mathcal{C}|}$. Furthermore, to enable logical negation of concepts, we augment the predicate space as:
\begin{equation}
    \mathbf{p} = [\bar{\mathbf{c}}; \neg\bar{\mathbf{c}}] \in \{0,1\}^{2|\mathcal{C}|}
\end{equation}
allowing rules involving both presence and absence of concepts (e.g., $\texttt{leg\_jump} \land \neg\texttt{leg\_static}$).

\subsubsection{Logical Layer Architecture}

We compose concept predicates into action-discriminative rules through stacked logical layers (Fig.~\ref{fig:model}(c)). Each layer $\mathcal{N}^{(l)}$ contains two sub-layers: conjunction (AND) and disjunction (OR), enabling arbitrary propositional formulas in disjunctive normal form~\cite{wang2023learning}. Let $\mathbf{n}^{(l-1)}$ denote the input (with $\mathbf{n}^{(0)} = \mathbf{p}$). The conjunction ($\mathcal{A}^{(l)}$) and disjunction layers ($\mathcal{B}^{(l)}$) compute:
\begin{equation}
    a_i^{(l)} = \bigwedge_{j: W_{ij}^{(\land)} = 1} n_j^{(l-1)}, \qquad
    b_i^{(l)} = \bigvee_{j: W_{ij}^{(\lor)} = 1} n_j^{(l-1)}
\end{equation}
where $\mathbf{W}^{(\land)}, \mathbf{W}^{(\lor)} \in \{0,1\}$ are learnable switchboards (i.e., binary adjacency matrices) specifying which predicates participate in each conjunction/disjunction. By stacking $L$ layers with outputs $\mathbf{n}^{(l)} = [\mathbf{a}^{(l)}; \mathbf{b}^{(l)}]$, the network learns compositional rules of increasing complexity, yielding rule activations $\mathbf{r} \in \{0,1\}^R$.

\subsubsection{Continuous Relaxation and Gradient Grafting}

Discrete Boolean operations in logical layer produce zero gradients. For end-to-end training, we adopt differentiable logic relaxations~\cite{barbiero2023interpretable} that approximate AND/OR via product-based formulations. During training, we maintain continuous weights $\tilde{\mathbf{W}}^{(\land)}, \tilde{\mathbf{W}}^{(\lor)} \in [0,1]$ and compute soft operations:
\begin{equation}
    \tilde{a}_i^{(l)} = \prod_{j} \left(1 - \tilde{W}_{ij}^{(\land)}(1 - \tilde{n}_j^{(l-1)})\right)
    \label{eq:soft_logic1}
\end{equation}
\begin{equation}
    \tilde{b}_i^{(l)} = 1 - \prod_{j} \left(1 - \tilde{W}_{ij}^{(\lor)} \tilde{n}_j^{(l-1)}\right)
    \label{eq:soft_logic2}
\end{equation}
When weights and inputs are binary, these reduce to standard AND/OR (Appendix~\ref{app:logic_layers}). Logical layer training employs gradient grafting~\cite{wang2023learning} (Fig~\ref{fig:model}(a)): the forward pass uses binarized concepts $\bar{\mathbf{c}}$ and discrete weights $\mathbf{W} = \mathbb{1}[\tilde{\mathbf{W}} > 0.5]$ for interpretable rule evaluation, while backpropagation computes gradients through the continuous relaxations (Eq.~\ref{eq:soft_logic1}and~\ref{eq:soft_logic2}) to update soft weights $\tilde{\mathbf{W}}$. For the concept binarizer, the Straight-Through Estimator~\cite{bengio2013estimating} passes gradients unchanged: $\partial \mathcal{L}/\partial \hat{c}_i \approx \partial \mathcal{L}/\partial \bar{c}_i$, as shown in Fig~\ref{fig:model}(a).

\subsection{Rule-Based Action Classification}
\label{sec:action_classification}

The rule activations $\mathbf{r} \in \{0,1\}^R$ encode which logical concept combinations are present. A linear classifier maps these to action predictions $\hat{\mathbf{y}} = \mathbf{V}\mathbf{r} + \mathbf{b}$, where $\mathbf{V} \in \mathbb{R}^{|\mathcal{A}| \times R}$, associates rules with actions, and $\mathbf{b} \in \mathbb{R}^{|\mathcal{A}|}$ is a bias term. Post-training, we extract human-readable rules by tracing adjacency matrices back to concept combinations, yielding expressions like $r_1: \texttt{leg\_jump} \land \lnot\texttt{arm\_swing} \land \texttt{motion\_upward}$. This ensures both \emph{local interpretability} (instance-specific concept activations) and \emph{global interpretability} (logical rules characterizing each action class). The details for post-training rule interpretation are provided in Appendix~\ref{app:logic_layers}.

\begin{table*}[!ht]
\centering
\caption{Comparison with state-of-the-art methods on NTU RGB+D 60/120, and NW-UCLA. \textbf{Cat.}: Categories. \textbf{Trans.}: Transformers \textbf{Mod.}: input modalities (J: Joint, B: Bone, JM: Joint Motion, BM: Bone Motion). \textbf{Two-stage}: requires separate pretraining. \textbf{Interpret.}: provides interpretable predictions. Best performance in \textbf{bold}, while second-best is \underline{underlined}.}
\label{tab:benchmark}
\small
\resizebox{0.9\textwidth}{!}{%
\begin{tabular}{llccccccccc}
\hline
\multirow{2}{*}{\textbf{Cat.}} & \multirow{2}{*}{\textbf{Methods}} & \multirow{2}{*}{\textbf{Venue}} & \multirow{2}{*}{\textbf{Mod.}} & \multicolumn{2}{c}{\textbf{NTU-RGB+D 60}} & \multicolumn{2}{c}{\textbf{NTU-RGB+D 120}} & \multirow{2}{*}{\textbf{NW-UCLA}} & \multirow{2}{*}{\textbf{Two-stage}} & \multirow{2}{*}{\textbf{Interpret.}} \\ \cline{5-8}
 &  &  &  & \textbf{X-Sub} & \textbf{X-View} & \textbf{X-Sub} & \textbf{X-Set} &  &  &  \\ \hline
 & ST-GCN & AAAI'18 & J+B & 81.5 & 88.3 & 70.7 & 73.2 & -- & \cellcolor{teal!30}N & \cellcolor{red!30}N \\
 & 2s-AGCN & CVPR'19 & J+B & 88.5 & 95.1 & 82.5 & 84.2 & -- & \cellcolor{teal!30}N & \cellcolor{red!30}N \\
 & MS-G3D & CVPR'20 & J+B & 86.0 & 94.1 & 80.2 & 86.1 & -- & \cellcolor{teal!30}N & \cellcolor{red!30}N \\
 & UNIK & BMVC'21 & J+B & 86.8 & 94.4 & 80.8 & 86.5 & -- & \cellcolor{teal!30}N & \cellcolor{red!30}N \\
 & InfoGCN & CVPR'20 & J+B & -- & -- & 88.5 & 89.7 & -- & \cellcolor{teal!30}N & \cellcolor{red!30}N \\
\rowcolor{gray!10} & \textbf{REASON (Ours)} & This work & J+B & 89.4 & 95.6 & 88.97 & 89.2 & -- & \cellcolor{teal!30}N & \cellcolor{green!30}Y \\ \cline{2-11} 
 & DC-GCN+ADG & ECCV'20 & J+B+JM+BM & 90.8 & 96.6 & 86.5 & 88.1 & 95.3 & \cellcolor{teal!30}N & \cellcolor{red!30}N \\
 & MS-G3D & CVPR'20 & J+B+JM+BM & 91.5 & 96.2 & 86.9 & 88.4 & -- & \cellcolor{teal!30}N & \cellcolor{red!30}N \\
 & MST-GCN & Access'21 & J+B+JM+BM & 91.5 & 96.6 & 87.5 & 88.8 & -- & \cellcolor{teal!30}N & \cellcolor{red!30}N \\
 & CTR-GCN & ICCV'21 & J+B+JM+BM & 92.4 & 96.4 & 88.9 & 90.6 & 96.5 & \cellcolor{teal!30}N & \cellcolor{red!30}N \\
GCN & EfficientGCN-B4 & TPAMI'22 & J+B+JM+BM & 91.7 & 95.7 & 88.3 & 89.1 & -- & \cellcolor{teal!30}N & \cellcolor{red!30}N \\
 & InfoGCN & CVPR'20 & J+B+JM+BM & 92.7 & 96.9 & 89.4 & 90.7 & 96.6 & \cellcolor{teal!30}N & \cellcolor{red!30}N \\
 & FR Head & CVPR'23 & J+B+JM+BM & 92.8 & 96.8 & 89.5 & 90.9 & 96.8 & \cellcolor{teal!30}N & \cellcolor{red!30}N \\
 & HD-GCN* & ICCV'23 & J+B+J'+B' & 93.0 & 97.0 & 89.8 & 91.2 & 96.9 & \cellcolor{teal!30}N & \cellcolor{red!30}N \\
 & DS-GCN & AAAI'24 & J+B+JM+BM & 93.1 & 97.5 & 89.2 & 90.3 & -- & \cellcolor{teal!30}N & \cellcolor{red!30}N \\
 & BlockGCN & CVPR'24 & J+B+JM+BM & 93.1 & 97.0 & 90.3 & 91.5 & 96.9 & \cellcolor{teal!30}N & \cellcolor{red!30}N \\
 & Hyper-GNN & TIP'21 & J+B+JM+BM & 89.5 & 95.7 & -- & -- & -- & \cellcolor{teal!30}N & \cellcolor{red!30}N \\
 & Selective-HCN & ICMR'22 & J+B+JM+BM & 90.8 & 96.6 & -- & -- & -- & \cellcolor{teal!30}N & \cellcolor{red!30}N \\
 & DST-HCN & ICME'23 & J+B+JM+BM & 92.3 & 96.8 & 88.8 & 90.7 & 96.6 & \cellcolor{teal!30}N & \cellcolor{red!30}N \\
 & Hyper-GCN & ICCV'25 & J+B+JM+BM & 93.7 & 97.8 & \underline{90.9} & 92.0 & 97.6 & \cellcolor{teal!30}N & \cellcolor{red!30}N \\
\rowcolor{gray!10} & \textbf{REASON (Ours)} & This work & J+B+JM+BM & 94.3 & 97.8 & 90.1 & \underline{91.9} & 97.9 & \cellcolor{teal!30}N & \cellcolor{green!30}Y \\ \hline
 & DSTA-Net & ACCV'20 & J+B+JM+BM & 89.5 & 95.7 & 86.6 & 89.0 & -- & \cellcolor{teal!30}N & \cellcolor{red!30}N \\
Trans. & IIP-Transformer & DATA'23 & J+B+JM+BM & 89.5 & 95.7 & 89.9 & 90.9 & -- & \cellcolor{teal!30}N & \cellcolor{red!30}N \\
 & SkateFormer & ECCV'24 & J+B+JM+BM & 93.5 & 97.8 & 89.8 & 91.4 & \textbf{98.3} & \cellcolor{teal!30}N & \cellcolor{red!30}N \\
 & USDRL & TPAMI'25 & J+M+B & 87.1 & 93.2 & 79.3 & 80.6 & -- & \cellcolor{orange!30}Y & \cellcolor{red!30}N \\ \hline
\multirow{3}{*}{LLM} & GAP & ICCV'23 & J+B+JM+BM & 92.9 & 97.0 & 89.9 & 91.1 & 97.2 & \cellcolor{teal!30}N & \cellcolor{red!30}N \\
 & LLM-AR & CVPR'24 & J & 95.0 & \textbf{98.4} & 88.7 & 91.5 & -- & \cellcolor{orange!30}Y & \cellcolor{green!30}Y \\
 & SUGAR & AAAI'26 & J & \underline{95.2} & 97.8 & 90.1 & 89.7 & -- & \cellcolor{orange!30}Y & \cellcolor{green!30}Y \\
\rowcolor{gray!10} & \textbf{REASON+LLM (Ours)} & This work & J & \textbf{95.6} & \underline{98.3} & \textbf{91.2} & \textbf{92.8} & \underline{98.1} & \cellcolor{teal!30}N & \cellcolor{green!30}Y \\ \hline
\end{tabular}
}
\end{table*}

\noindent\textbf{Overall Training Objective}
\label{sec:training}
The framework is trained end-to-end by combining all loss terms: $\mathcal{L} = \mathcal{L}_\text{task} + \alpha \mathcal{L}_\text{concept} + \beta \mathcal{L}_\text{align} + \gamma \mathcal{L}_\text{div} + \lambda \|\tilde{\mathbf{W}}\|_1$, where the $\ell_1$ regularization on $\tilde{\mathbf{W}}$ enforces sparse, interpretable rules. At inference, concepts are binarized, rules are evaluated discretely, and actions are predicted via triggered rules.



\section{Experiments}
\label{sec:experiments}
We evaluate our neurosymbolic framework on standard skeleton-based HAR benchmarks, focusing on: (i) performance competitiveness, (ii) interpretability of learned rules, and (iii) component-wise contributions.

\subsection{Datasets}

We evaluate on three benchmarks: \textbf{NTU RGB+D 60} (56,880 sequences, 60 classes) under Cross-Subject (X-Sub) and Cross-View (X-View) protocols; \textbf{NTU RGB+D 120} (114,480 sequences, 120 classes) under Cross-Subject (X-Sub) and Cross-Setup (X-Set) protocols; and \textbf{NW-UCLA} (1,494 sequences, 10 classes) following protocol to train on cameras 1\&2, test on camera 3.

\subsection{Implementation Details}

We follow~\cite{chen2021channel} for preprocessing, using random rotation, spatial flip, and temporal cropping with uniform 64-frame sampling. Training uses AdamW for 400 epochs with cosine decay on a single A6000 GPU.

We employ \emph{staggered warmup} to prevent spurious rule formation: the GCN/CLIP/STC-Decoder warm up for 5 epochs while logic layers remain frozen for 15 epochs. This delay ensures logic layers learn from converged concept predictions rather than random activations. Logic layers use $10\times$ higher learning rate ($10^{-4}$ vs. $10^{-5}$) to traverse from dense initialization to sparse binary connectivity against $\ell_1$ regularization. Full hyperparameters are in Appendix~\ref{app:impl_details}.





\subsection{Comparison with State-of-the-Art}

Table~\ref{tab:benchmark} compares against GCN-based, Transformer-based, and LLM-augmented methods. Beyond performance, we highlight whether methods require \emph{two-stage} training (separate pretraining) and whether they provide \emph{interpretable} predictions.

Our core neurosymbolic framework (\textsc{REASON}) achieves competitive performance while being the only single-stage, fully interpretable approach. In 4-stream configuration, we attain 94.3\% on NTU-60 X-Sub and 92.8\% on NTU-120 X-Set, surpassing black-box methods (Hyper-GCN, BlockGCN). Critically, the concept-rule reasoning chain remains fully auditable, distinguishing our approach from prior work.

\noindent\textbf{LLM Re-ranking (Optional).} To enable fair comparison with LLM-augmented methods (SUGAR, LLM-AR), we evaluate an optional re-ranking stage. Unlike these methods that use generic action descriptions, we ground a LoRA-finetuned LLaMA-3.2 7B in \emph{instance-specific} concept activations and top-$K$ classifier candidates (Appendix~\ref{app:llm_reranking}). With LLM re-ranking, we achieve 95.6\% (X-Sub) and 93.2\% (X-Set), outperforming two-stage LLM methods, as we reason over interpretable concepts rather than opaque features.

\begin{table*}[!t]
\centering
\caption{Ablation studies of different components of our proposed framework on NTU-120.
(a) Backbone ablation.
(b) Component ablation.
(c) Group query configuration.
(d) Concept vocabulary composition.
(e) Logic layer capacity analysis.
(f) Concept Intervention Analysis.}
\vspace*{-5mm}
\label{tab:ablations_grid}

\setlength{\tabcolsep}{3pt}

\begin{tabular}{@{}c@{\hspace{0.8em}}c@{\hspace{0.8em}}c@{}}

\begin{minipage}[t]{0.3\textwidth}\vspace{0pt}\centering\small
(a) Backbone ablation\par\vspace{0.1em}
\begin{adjustbox}{max width=\linewidth}
\begin{tabular}{lccc}
\toprule
\multirow{2}{*}{\textbf{Backbone}} & \multicolumn{3}{c}{\textbf{Accuracy (\%)}} \\
\cmidrule(lr){2-4}
 & w/o $\mathcal{L}_{\text{align}}$ & w/o $\mathcal{L}_{\text{div}}$ & w/ both \\
\midrule
ST-GCN & 80.07 & 81.15 & 82.40 \\
2s-AGCN & 82.15 & 83.20 & 84.85 \\
CTR-GCN & 82.90 & 84.10 & 86.12 \\
ST-GCN++ & 83.25 & 84.65 & 86.70 \\
Hyper-GCN & 84.08 & 85.27 & \textbf{87.34} \\
\bottomrule
\end{tabular}
\end{adjustbox}
\end{minipage}
&
\begin{minipage}[t]{0.3\textwidth}\vspace{0pt}\centering\small
(c) Group query configuration\par\vspace{0.1em}
\begin{adjustbox}{max width=\linewidth}
\begin{tabular}{ccccc}
\hline
\multirow{2}{*}{\textbf{$G_\text{s}$}} & \multirow{2}{*}{\textbf{$G_\text{t}$}} & \multirow{2}{*}{\textbf{Total $G$}} & \multicolumn{2}{c}{\textbf{Acc. (\%)}} \\ \cline{4-5} 
 &  &  & \textbf{X-sub} & \textbf{X-set} \\ \hline
\multicolumn{3}{l}{Baseline (simple FC layer)} & 86.20 & 85.70 \\ \hline
4 & 2 & 6 & 85.72 & 85.12 \\
8 & 4 & 12 & \textbf{86.38} & \textbf{87.34} \\
16 & 8 & 24 & 86.11 & 87.21 \\
32 & 16 & 48 & 85.81 & 87.18 \\
51 & 16 & 67 & 86.03 & 87.09 \\ \hline
\end{tabular}
\end{adjustbox}
\end{minipage}
&
\begin{minipage}[t]{0.3\textwidth}\vspace{0pt}\centering\small
(e) Logic layer capacity\par\vspace{0.1em}
\begin{adjustbox}{max width=\linewidth}
\begin{tabular}{lccccc}
\toprule
Config & NOT & Skip & L1/L2 & Final & Acc. \\
\midrule
Small & \cmark & \cmark & 64 & 390 & 83.60 \\
Medium & \cmark & \cmark & 128 & 646 & \textbf{87.34} \\
Large & \cmark & \cmark & 256 & 1158 & 86.18 \\
XLarge & \cmark & \cmark & 512 & 2182 & 81.38 \\
\midrule
No Skip & \cmark & \xmark & 128 & 256 & 83.77 \\
No NOT & \xmark & \cmark & 128 & 579 & 49.87 \\
\bottomrule
\end{tabular}
\end{adjustbox}
\end{minipage}
\\[0.8em]

\begin{minipage}[t]{0.3\textwidth}\vspace{0pt}\centering\small
(b) Component ablation\par\vspace{0.1em}
\begin{adjustbox}{max width=\linewidth}
\begin{tabular}{lcc}
\toprule
\textbf{Model Configuration} & \textbf{X-Sub} & \textbf{X-Set} \\
\midrule
Hyper-GCN & 84.28 & 84.08 \\
+ Text Alignment ($\mathcal{L}_\text{align}$) & 85.20 & 85.27 \\
+ Part Divergence ($\mathcal{L}_\text{div}$) & 86.20 & 85.70 \\
+ STC-Decoder & 86.38 & 87.34 \\
\bottomrule
\end{tabular}
\end{adjustbox}
\end{minipage}
&
\begin{minipage}[t]{0.3\textwidth}\vspace{0pt}\centering\small
(d) Concept vocabulary composition\par\vspace{0.1em}
\begin{adjustbox}{max width=\linewidth}
\begin{tabular}{lcc}
\toprule
\textbf{Concept Configuration} & \# Concepts & Accuracy \\
\midrule
Spatial only & 51 & 78.39 \\
Spatial + Temporal & 66 & 86.15 \\
+ Interaction concepts & 67 & 87.34 \\
\bottomrule
\end{tabular}
\end{adjustbox}
\end{minipage}
&
\begin{minipage}[t]{0.3\textwidth}\vspace{0pt}\centering\small
(f) Concept intervention\par\vspace{0.1em}
\begin{adjustbox}{max width=\linewidth}
\begin{tabular}{lcc}
\toprule
\textbf{Intervention Level} & \textbf{Accuracy (\%)} & \textbf{$\Delta$ Acc.} \\
\midrule
No intervention & 87.35 & -- \\
Correct 1 concept & 89.23 & +1.88 \\
Correct 2 concepts & 91.21 & +3.86 \\
Correct 3 concepts & 93.02 & +5.67 \\
\bottomrule
\end{tabular}
\end{adjustbox}
\end{minipage}

\end{tabular}
\end{table*}

\subsection{Interpretable Reasoning via Concept-Rule Chains}

\begin{figure}[!ht]
    \centering
    \includegraphics[width=\linewidth]{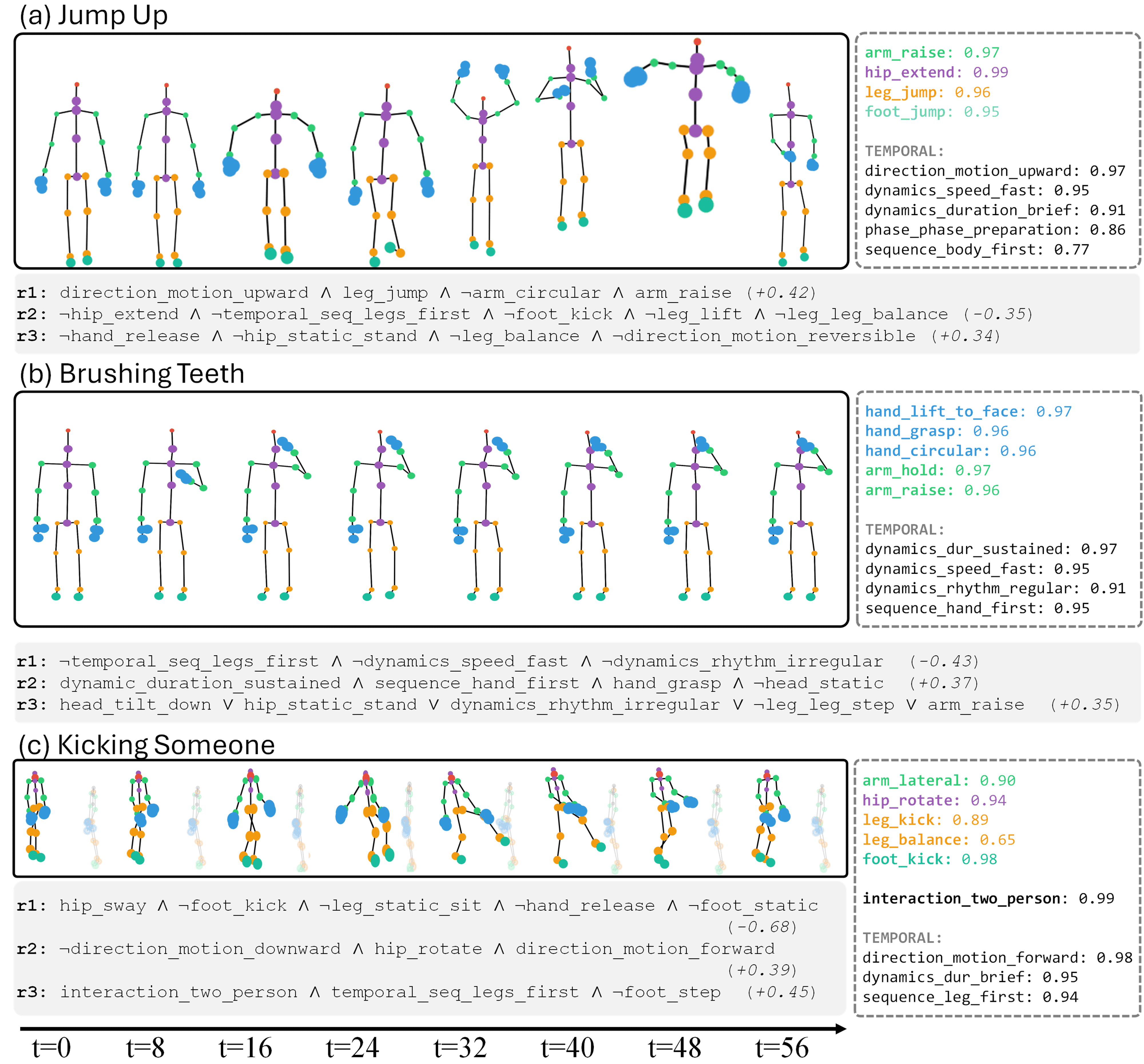}
    \caption{\textbf{Visualization of our concept-grounded neurosymbolic reasoning.} Each panel shows temporally sampled skeletons with concept activations, top spatial–temporal concepts (dotted box), and top-3 learned logical rules with weights (gray box).}
    \label{fig:skeleton}
\end{figure}

Beyond accuracy, our framework provides transparent reasoning through learned concept-rule chains. Figure~\ref{fig:skeleton} visualizes the complete reasoning pipeline for representative actions. Skeleton poses are rendered with joint sizes scaled by concept activation confidence, revealing which body parts drive predictions. For \textsc{BrushingTeeth}, the model activates intuitive concepts (\texttt{hand\_lift\_to\_face}, \texttt{duration\_sustained}, \texttt{rhythm\_regular}) and applies learned rules such as $r_1: \texttt{duration\_sustained} \land \texttt{hand\_grasp} \land \neg\texttt{head\_static}$ (+0.37). This end-to-end transparency, from skeleton input to concept activations to logical rule firings, enables systematic analysis of both correct classifications and failure cases, a capability absent in existing methods.

\subsection{Ablation Studies}
We conduct ablations on NTU RGB+D 120 (Table~\ref{tab:ablations_grid}).

\noindent\textbf{Backbone Generalization} (Table~\ref{tab:ablations_grid}a). Our framework is encoder-agnostic: across ST-GCN, 2s-AGCN, CTR-GCN, ST-GCN++, and Hyper-GCN, our approach consistently improves performance, demonstrating plug-and-play applicability for existing GCN encoders.

\noindent\textbf{Component Analysis} (Table~\ref{tab:ablations_grid}b). Each module contributes incrementally: text alignment ($\mathcal{L}_\text{align}$) adds +0.9-1.2\%, part divergence ($\mathcal{L}_\text{div}$) +0.4-1.0\%, and STC-Decoder +0.2--1.6\%. 

\noindent\textbf{STC-Decoder Configuration} (Table~\ref{tab:ablations_grid}c). Group-decoding with $G_\text{s}=8$, $G_\text{t}=4$ achieves optimal accuracy while reducing cross-attention cost by $\sim$6$\times$ compared to per-concept queries ($G=74$).

\noindent\textbf{Concept Vocabulary} (Table~\ref{tab:ablations_grid}d). Spatial concepts alone achieve 78.39\%; adding temporal concepts yields +7.8\% (86.15\%), demonstrating that motion dynamics are critical for distinguishing actions with similar poses. Interaction concepts provide a further +1.2\%.

\noindent\textbf{Logic Layer Capacity} (Table~\ref{tab:ablations_grid}e). The 128-node configuration is optimal; smaller (64) lacks expressiveness while larger (256/512) overfits. Crucially, disabling negation causes catastrophic failure (49.87\%), many actions require $\neg$concepts (e.g., \textsc{Standing} needs $\neg$\texttt{leg\_walk}). Skip connections contribute +3.6\% by preserving atomic concept access.

\noindent\textbf{Concept Intervention} (Table~\ref{tab:ablations_grid}f). Replacing mispredicted concepts $\hat{c}_i$ with ground-truth $c^*_i$ improves accuracy monotonically: +1.9\% (1 concept), +3.9\% (2), +5.7\% (3). This confirms that logic layers capture faithful concept–action relationships and errors stem from concept misprediction rather than flawed rules, enabling human-in-the-loop correction. Overall process of concept intervention is in Appendix~\ref{app:intervention}.

\begin{figure}[!h]
    \centering
    \includegraphics[width=0.7\linewidth]{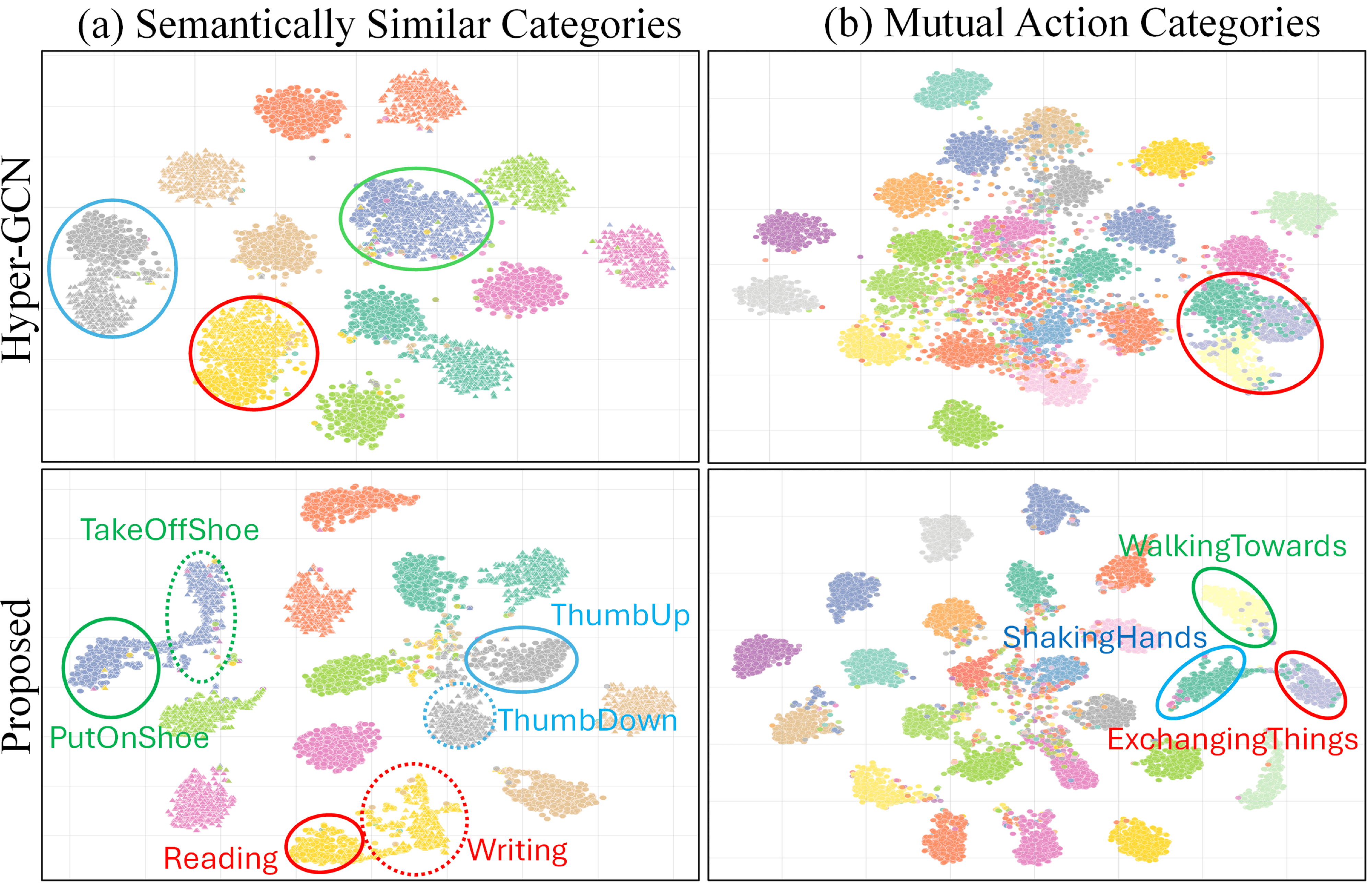}
    \caption{\textbf{t-SNE visualization on NTU RGB+D 120}. Hyper-GCN features (top) and our STC-decoder features (bottom). (a) Semantically similar action pairs. (b) Two-person interaction actions.}
    \label{fig:tsne}
\end{figure}

\noindent\textbf{Feature Space Analysis} Figure~\ref{fig:tsne} compares t-SNE embeddings of Hyper-GCN features and STC-Decoder concept activations. Actions overlapping in the Hyper-GCN space (e.g., \textsc{Reading}/\textsc{Writing}, \textsc{PutOn}/\textsc{TakeOffShoe}, \textsc{ThumbUp}/\textsc{Down}) are well separated in the concept space by discriminative temporal concepts (e.g., \texttt{motion\_forward} vs. \texttt{motion\_backward}), with similar separation for mutual actions (e.g., \textsc{ExchangingThings}, \textsc{ShakingHands}).

\section{Conclusion}

In this work, we presented \textsc{REASON}, a neurosymbolic framework that formulates skeleton-based HAR as logical reasoning over learnable motion concepts. Through concept grounding, spatio-temporal decoding, and differentiable logic, \textsc{REASON} achieves state-of-the-art performance while maintaining fully transparent concept-rule reasoning chains, establishing neurosymbolic learning as a promising paradigm for interpretable skeleton-based HAR.







\bibliographystyle{named}
\bibliography{ijcai26}

\clearpage
\appendix

\section*{Appendix}
\addcontentsline{toc}{section}{Appendix}

\section{Concept Bank Generation Details}
\label{app:concept_bank}

\begin{figure}[!ht]
    \centering
    \includegraphics[width=0.5\textwidth]{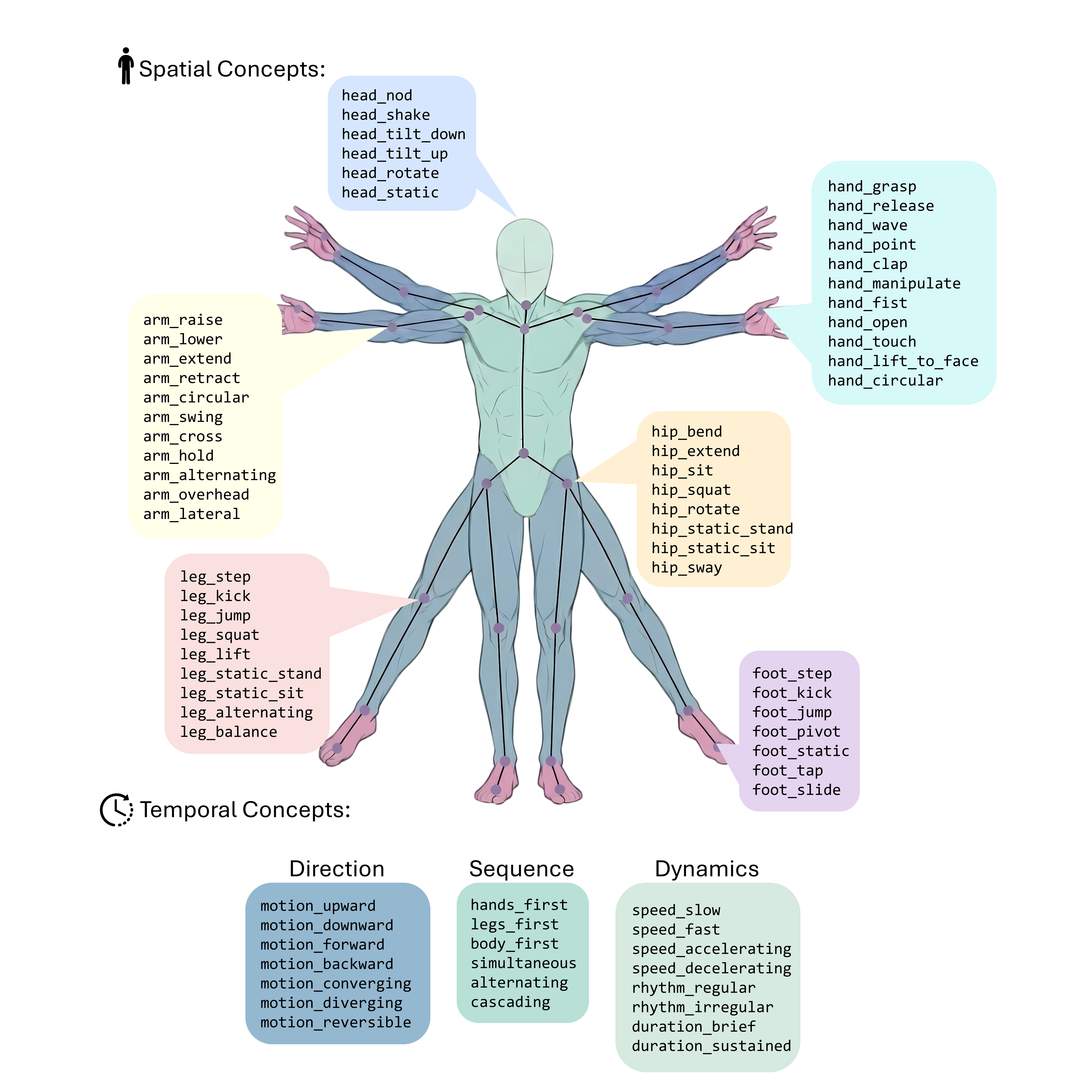}
    \caption{Generated concept bank for NTU-RGB+D 60/120 and NW-UCLA, visualizing spatial and temporal concepts.}
    \label{fig:venturi_man}
\end{figure}

This appendix provides implementation details for the concept bank construction pipeline described in Sec.~\ref{sec:concept_bank}. The final concept bank generated for NTU-RGB+D 60/120 and NW-UCLA is visualized in Figure~\ref{fig:venturi_man}.

\subsection{Stage 1: Action Description Prompts}

For each action $a$ in the dataset, we query GPT-3.5 with the following structured prompt:

\begin{promptbox}
\small
\textit{``Describe how each body part moves when a person performs [action]: head, hand, arm, hip, leg, foot.''}
\end{promptbox}

\noindent This yields fine-grained descriptions $\mathcal{D}_a = \{d_a^{(p)}\}_{p \in \mathcal{P}}$ for each body part $p$. For instance, for the action \textsc{Jump}:

\begin{outputbox}
\textit{Head}: tilts slightly upward; 
\textit{Hand}: swings forward and upward; 
\textit{Arm}: extends overhead; 
\textit{Hip}: flexes then rapidly extends; 
\textit{Leg}: bends at knees then pushes explosively; 
\textit{Foot}: pushes off ground, leaves surface.
\end{outputbox}

\noindent While these descriptions are action-specific and detailed, using these raw descriptions as concepts directly would cause a vocabulary explosion and prevent the model from learning shared logical structures across different actions.

\subsection{Stage 2: Clustering Details}

To obtain a compact, shared vocabulary, we abstract the fine-grained descriptions into canonical motion concepts. For each body part $p \in \mathcal{P}$, we we identify verb-direction-modifier tuples (e.g. \textit{extends\_forward}, \textit{bends\_slightly}) in the motion patterns from all the motion descriptions $\{d_a^{(p)}\}_{a \in \mathcal{A}}$. We then embed these patterns using a sentence encoder $\phi$ (we use all-MiniLM-L6-v2) and cluster them via $k$-means, determining the optimal $k$ by the elbow method and shown below.

\begin{equation}
    \{c_1^{(p)}, \ldots, c_K^{(p)}\} = \text{KMeans}\left(\{\phi(m_i^{(p)})\}_{i=1}^N, k^*\right)
\end{equation}

where $\{m_1^{(p)}, \ldots, m_N^{(p)}\}$ are the extracted patterns for each body part $p$. The most frequent pattern within each cluster serves as the representative concept, which yields spatial concepts such as \texttt{arm\_raise}, \texttt{leg\_jump}, \texttt{hand\_grasp} and alike.

\subsection{Temporal Concept Categories}

Beyond spatial concepts derived from clustering, we define a set of \emph{temporal concepts} that capture motion dynamics across time. These concepts abstract how body parts move, evolve, and coordinate during an action, and are visualized in Figure~\ref{fig:venturi_man} and summarized in Table~\ref{tab:concept_stats}.

Specifically, temporal concepts are organized into three categories:
\begin{itemize}
    \item \textbf{Direction}, describing dominant motion trends such as \texttt{motion\_upward}, \texttt{motion\_downward}, \texttt{motion\_forward}, \texttt{motion\_backward}, as well as relational patterns including \texttt{converging} and \texttt{diverging}.
    \item \textbf{Sequence}, characterizing the ordering and coordination of body-part activations, e.g., \texttt{hands\_first}, \texttt{legs\_first}, \texttt{body\_first}, \texttt{simultaneous}, and \texttt{alternating}.
    \item \textbf{Dynamics}, capturing temporal evolution properties such as \texttt{speed\_slow}, \texttt{speed\_fast}, \texttt{accelerating}, \texttt{decelerating}, \texttt{rhythm\_regular}, \texttt{duration\_brief}, and \texttt{duration\_sustained}.
\end{itemize}

Together, these temporal concepts complement spatial descriptors by encoding \emph{how} movements unfold over time, enabling the model to distinguish actions that are spatially similar but temporally distinct.

\subsection{Stage 3: Concept Matching Prompts}
Given the concept vocabulary $\mathcal{C}$, we construct an action-concept association matrix $\mathbf{M} \in \{0, 1\}^{|\mathcal{A}| \times |\mathcal{C}|}$, where $\mathbf{M}_{a,c} = 1$ indicates that concept $c$ is active for action $a$. Crucially, we allow \emph{multiple concepts per body part} to capture complex movements, for example, \textsc{Jump} activates both \texttt{leg\_squat} (preparatory crouch) and \texttt{leg\_jump} (explosive extension) for the leg.

To populate the action-concept matrix $\mathbf{M}$, we employ LLaMA-3.2 7B with the following prompt for each action $a$ and body part $p$:

\begin{promptbox}
\small
\textit{``Given the movement `[description]' for [body part] during [action], select 1--3 concepts from: [concept list] that best describe this movement.''}
\end{promptbox}

\noindent The model returns a ranked list of concepts, which we threshold to obtain binary activations. This enables multi-concept activation where compositional movements trigger multiple predicates.

\subsection{Duplicate Resolution}

When multiple actions share identical concept vectors (e.g., semantically similar actions like \textsc{WearGlasses} and \textsc{TakeOffGlasses}), we invoke the LLM with:

\begin{promptbox}
\small
\textit{``Actions [action1] and [action2] have similar concept activations. What distinguishes them? Select differentiating concepts from: [concept list].''}
\end{promptbox}

\noindent This refinement typically identifies opposing temporal direction concepts (\texttt{motion\_forward} vs. \texttt{motion\_backward}) or complementary spatial concepts (\texttt{arm\_raise} vs. \texttt{arm\_lower}).

\subsection{Algorithm Summary}

Algorithm~\ref{alg:concept_bank} summarizes the complete concept bank construction pipeline.

\begin{algorithm}[h]
\caption{Concept Bank Construction}
\label{alg:concept_bank}
\begin{algorithmic}[1]
\REQUIRE Action set $\mathcal{A}$, body parts $\mathcal{P}$, LLM $\mathcal{M}_\text{large}$, LLM $\mathcal{M}_\text{small}$
\ENSURE Concept vocabulary $\mathcal{C}$, association matrix $\mathbf{M}$
\STATE \textcolor{gray}{\textit{// Stage 1: Generate action descriptions}}
\FOR{each action $a \in \mathcal{A}$}
    \STATE $\mathcal{D}_a \gets \mathcal{M}_\text{large}(\texttt{describe\_parts}(a, \mathcal{P}))$
\ENDFOR
\STATE \textcolor{gray}{\textit{// Stage 2: Cluster into abstract concepts}}
\FOR{each body part $p \in \mathcal{P}$}
    \STATE Extract patterns $\{m_i^{(p)}\}$ from $\{\mathcal{D}_a\}$
    \STATE $k^* \gets \texttt{ElbowMethod}(\{\phi(m_i^{(p)})\})$
    \STATE $\mathcal{C}_p \gets \texttt{KMeans}(\{\phi(m_i^{(p)})\}, k^*)$
\ENDFOR
\STATE $\mathcal{C}_\text{spatial} \gets \bigcup_{p \in \mathcal{P}} \mathcal{C}_p$
\STATE $\mathcal{C} \gets \mathcal{C}_\text{spatial} \cup \mathcal{C}_\text{temporal} \cup \mathcal{C}_\text{interaction}$
\STATE \textcolor{gray}{\textit{// Stage 3: Associate concepts to actions}}
\FOR{each action $a \in \mathcal{A}$}
    \FOR{each part $p \in \mathcal{P}$}
        \STATE $\mathbf{M}_{a, \mathcal{C}_p} \gets \mathcal{M}_\text{small}(\texttt{match}(d_a^{(p)}, \mathcal{C}_p))$
    \ENDFOR
    \STATE $\mathbf{M}_{a, \mathcal{C}_\text{temp}} \gets \mathcal{M}_\text{small}(\texttt{classify\_temporal}(a))$
\ENDFOR
\STATE \textcolor{gray}{\textit{// Refinement: ensure unique signatures}}
\STATE $\mathbf{M} \gets \texttt{ResolveDuplicates}(\mathbf{M}, \mathcal{M}_\text{small})$
\RETURN $\mathcal{C}$, $\mathbf{M}$
\end{algorithmic}
\end{algorithm}

\subsection{Concept Bank Statistics}

Table~\ref{tab:concept_stats} summarizes the concept bank composition. The complete list of concepts is provided in Figure~\ref{fig:venturi_man}.

\begin{table}[h]
\centering
\caption{Concept bank statistics for NTU RGB+D and NW-UCLA.}
\label{tab:concept_stats}
\small
\begin{tabular}{lcc}
\toprule
\textbf{Category} & \textbf{Subcategory} & \textbf{\#Concepts} \\
\midrule
\multirow{6}{*}{Spatial ($\mathcal{C}_\text{s}$)} & Head & 6 \\
& Hand & 11 \\
& Arm & 11 \\
& Hip & 8 \\
& Leg & 9 \\
& Foot & 7 \\
\midrule
\multirow{3}{*}{Temporal ($\mathcal{C}_\text{t}$)} & Direction & 7 \\
& Sequence & 6 \\
& Dynamics & 8 \\
\midrule
Interaction ($\mathcal{C}_\text{int}$) & Two-person & 1 \\
\midrule
\textbf{Total} & & \textbf{74} \\
\bottomrule
\end{tabular}
\end{table}

\section{Skeleton Representation Learning Details}
\label{app:skeleton_rep}

\subsection{GCN Encoder Architecture}

We model the skeleton as a spatio-temporal graph $\mathcal{G} = (\mathcal{V}, \mathcal{E})$, where nodes $\mathcal{V}$ correspond to body joints and edges $\mathcal{E}$ encode anatomical connectivity. At layer $l$, node features $\mathbf{H}^{(l)} \in \mathbb{R}^{V \times D}$ are updated via graph convolution:
\begin{equation}
    \mathbf{H}^{(l+1)} = \sigma\left(\tilde{\mathbf{D}}^{-\frac{1}{2}} \tilde{\mathbf{A}} \tilde{\mathbf{D}}^{-\frac{1}{2}} \mathbf{H}^{(l)} \mathbf{W}^{(l)}\right)
\end{equation}
where $\tilde{\mathbf{A}} = \mathbf{A} + \mathbf{I}$ is the adjacency matrix with self-loops, $\tilde{\mathbf{D}}$ is the degree matrix, $\mathbf{W}^{(l)}$ are learnable weights, and $\sigma$ is a non-linearity. We stack $L$ layers with multi-scale temporal convolutions~\cite{chen2021channel} to capture motion dynamics across varying temporal receptive fields.

Our framework is agnostic to the specific GCN architecture; we experiment with CTR-GCN~\cite{chen2021channel} and Hyper-GCN~\cite{zhou2025adaptive} as backbones. The key requirement is retaining full spatio-temporal resolution (no global pooling) to preserve fine-grained information for concept prediction.

\subsection{LoRA Configuration for CLIP}

We fine-tune the CLIP text encoder using LoRA~\cite{hu2022lora} with rank $r=8$ and scaling factor $\alpha=16$. LoRA introduces trainable low-rank matrices $\mathbf{B}\mathbf{A}$ (where $\mathbf{A} \in \mathbb{R}^{r \times d}$, $\mathbf{B} \in \mathbb{R}^{d \times r}$) into the attention layers:
\begin{equation}
    \mathbf{W}' = \mathbf{W} + \frac{\alpha}{r} \mathbf{B}\mathbf{A}
\end{equation}
where $\mathbf{W}$ denotes frozen pretrained weights. This adaptation enables learning motion-specific semantics while preserving CLIP's general language understanding with minimal additional parameters ($<$1\% of total).

\begin{table}[h]
\centering
\small
\begin{tabular}{l c}
\toprule
\textbf{LoRA Setting} & \textbf{Value} \\
\midrule
Backbone & CLIP ViT-B/32 \\
Adapted Encoder & Text encoder only \\
LoRA Rank ($r$) & 16 \\
Scaling Factor ($\alpha$) & 32 \\
Target Parameters & Query ($q$), Key ($k$) \\
Injection Position & All transformer layers \\
Dropout Rate & 0.1 \\
Trainable Parameters & $0.39M$ \\
\bottomrule
\end{tabular}
\caption{LoRA configuration used to fine-tune the CLIP text encoder. The vision encoder is frozen and discarded during adaptation.}
\label{tab:lora_config}
\end{table}

\subsection{Discussion: Prompt Learning Alternatives}

Prompt learning approaches could alternatively learn continuous prompt vectors. However, such methods yield embeddings that lack direct interpretability, a property essential for our neurosymbolic framework where the alignment should reinforce human-understandable motion semantics. Our concept-grounded prompts explicitly encode compositional motion primitives, establishing semantic correspondence between language and skeleton modalities.

\section{STC-Decoder Implementation Details}
\label{app:stc_decoder}

\subsection{Cross-Attention Computation}

Each branch refines its group queries by attending to the decoupled skeleton features. The spatial branch computes:
\begin{equation}
    \text{CrossAttn}(\mathbf{Q}_\text{s}, \mathbf{F}_\text{s}) = \text{softmax}\left(\frac{\mathbf{Q}_\text{s} \mathbf{F}_\text{s}^\top}{\sqrt{D}}\right) \mathbf{F}_\text{s}
\end{equation}
followed by LayerNorm, residual connections, and a two-layer FFN. The temporal branch processes $\mathbf{Q}_\text{t}$ and $\mathbf{F}_\text{t}$ analogously. This allows each group query to selectively aggregate joint-level (spatial) or frame-level (temporal) information relevant to its concept subset.

\subsection{Group-FC Indexing Details}

Let $g_\text{s} = \lceil |\mathcal{C}_\text{s}| / G_\text{s} \rceil$ denote the \emph{group factor} (concepts per spatial group). For each group $k \in \{0, \ldots, G_\text{s}-1\}$, we maintain a learnable projection $\mathbf{W}_k^\text{s} \in \mathbb{R}^{D \times g_\text{s}}$. The logit for the $i$-th spatial concept is:
\begin{equation}
    \ell_i^\text{s} = \tilde{\mathbf{Q}}_\text{s}[k,:] \cdot \mathbf{W}_k^\text{s}[:,j], \quad \text{where } k = \lfloor i / g_\text{s} \rfloor, \; j = i \mod g_\text{s}
\end{equation}
This operation simultaneously expands each group query to $g_\text{s}$ outputs and projects from embedding dimension $D$ to scalar logits. The temporal branch follows identically with $G_\text{t}$ groups and $g_\text{t} = \lceil |\mathcal{C}_\text{t}| / G_\text{t} \rceil$.

\subsection{PyTorch-Style Implementation}

Algorithm~\ref{alg:stc_decoder} provides the complete STC-Decoder forward pass for spatio-temporal concept prediction.

\begin{algorithm}[h]
\caption{Spatio-Temporal Concept Decoder Forward Pass}
\label{alg:stc_decoder}
\small
\begin{algorithmic}[1]
\REQUIRE Encoder features $\mathbf{F} \in \mathbb{R}^{B \times T \times V \times D}$
\REQUIRE Learnable queries $\mathbf{Q}_\text{s} \in \mathbb{R}^{G_\text{s} \times D}$, $\mathbf{Q}_\text{t} \in \mathbb{R}^{G_\text{t} \times D}$
\REQUIRE Group-FC weights $\{\mathbf{W}_k^\text{s}\}_{k=0}^{G_\text{s}-1}$, $\{\mathbf{W}_k^\text{t}\}_{k=0}^{G_\text{t}-1}$
\ENSURE Concept predictions $\hat{\mathbf{c}} \in [0,1]^{B \times |\mathcal{C}|}$
\STATE
\STATE \textcolor{gray}{\textit{// Step 1: Spatio-temporal feature decoupling}}
\STATE $\mathbf{F}_\text{s} \gets \text{mean}(\mathbf{F}, \text{dim}=1)$ \textcolor{gray}{\textit{// [B, V, D] -- aggregate over time}}
\STATE $\mathbf{F}_\text{t} \gets \text{mean}(\mathbf{F}, \text{dim}=2)$ \textcolor{gray}{\textit{// [B, T, D] -- aggregate over joints}}
\STATE
\STATE \textcolor{gray}{\textit{// Step 2: Cross-attention with group queries}}
\STATE $\tilde{\mathbf{Q}}_\text{s} \gets \text{FFN}(\text{CrossAttn}(\mathbf{Q}_\text{s}, \mathbf{F}_\text{s}))$ \textcolor{gray}{\textit{// [B, $G_\text{s}$, D]}}
\STATE $\tilde{\mathbf{Q}}_\text{t} \gets \text{FFN}(\text{CrossAttn}(\mathbf{Q}_\text{t}, \mathbf{F}_\text{t}))$ \textcolor{gray}{\textit{// [B, $G_\text{t}$, D]}}
\STATE
\STATE \textcolor{gray}{\textit{// Step 3: Group-FC for spatial concepts}}
\STATE $\boldsymbol{\ell}_\text{s} \gets \text{empty}(B, |\mathcal{C}_\text{s}|)$
\FOR{$k = 0$ to $G_\text{s} - 1$}
    \STATE $\mathbf{g}_k \gets \tilde{\mathbf{Q}}_\text{s}[:, k, :]$ \textcolor{gray}{\textit{// [B, D] -- k-th group query}}
    \STATE $\boldsymbol{\ell}_k \gets \mathbf{g}_k \cdot \mathbf{W}_k^\text{s}$ \textcolor{gray}{\textit{// [B, $g_\text{s}$] -- project to concept logits}}
    \STATE $\boldsymbol{\ell}_\text{s}[:, k \cdot g_\text{s} : (k+1) \cdot g_\text{s}] \gets \boldsymbol{\ell}_k$
\ENDFOR
\STATE $\boldsymbol{\ell}_\text{s} \gets \boldsymbol{\ell}_\text{s}[:, :|\mathcal{C}_\text{s}|]$ \textcolor{gray}{\textit{// Trim to exact concept count}}
\STATE
\STATE \textcolor{gray}{\textit{// Step 4: Group-FC for temporal concepts (analogous)}}
\STATE $\boldsymbol{\ell}_\text{t} \gets \text{GroupFC}(\tilde{\mathbf{Q}}_\text{t}, \{\mathbf{W}_k^\text{t}\}, G_\text{t}, |\mathcal{C}_\text{t}|)$
\STATE
\STATE \textcolor{gray}{\textit{// Step 5: Concatenate and apply sigmoid}}
\STATE $\hat{\mathbf{c}} \gets \sigma([\boldsymbol{\ell}_\text{s}; \boldsymbol{\ell}_\text{t}])$ \textcolor{gray}{\textit{// [B, $|\mathcal{C}|$] -- soft concept activations}}
\STATE
\RETURN $\hat{\mathbf{c}}$
\end{algorithmic}
\end{algorithm}

\noindent\textbf{Implementation Note.}
In practice, the Group-FC loop (lines 12-16) is parallelized using
\texttt{torch.einsum} or batched matrix multiplication for efficiency:
\begin{lstlisting}[language=Python]
# W_s: [G_s, D, g_s], Q_s: [B, G_s, D]
logits_s = torch.einsum(
    'bgd,gdp->bgp', Q_s, W_s
)  # [B, G_s, g_s]
logits_s = logits_s.reshape(B, -1)[:, :num_spatial_concepts]
\end{lstlisting}

\subsection{Default Hyperparameters}
Ablation studies (Table~\ref{tab:ablations_grid}c) confirm that the default configuration ($G_\text{s}=8$, $G_\text{t}=4$) achieves optimal performance, outperforming both fewer groups (insufficient specialization) and per-concept queries (no grouping benefit).

\begin{table}[h]
\centering
\caption{STC-Decoder hyperparameters.}
\label{tab:stc_params}
\small
\begin{tabular}{lc}
\toprule
\textbf{Parameter} & \textbf{Value} \\
\midrule
Spatial groups $G_\text{s}$ & 8 \\
Temporal groups $G_\text{t}$ & 4 \\
Spatial concepts $|\mathcal{C}_\text{s}|$ & 51 \\
Temporal concepts $|\mathcal{C}_\text{t}|$ & 15 \\
Interaction concepts $|\mathcal{C}_\text{int}|$ & 1 \\
Embedding dimension $D$ & 256 \\
FFN hidden dimension & 512 \\
Number of attention heads & 4 \\
Dropout & 0.1 \\
\bottomrule
\end{tabular}
\end{table}




\section{Differentiable Logic Layer Details}
\label{app:logic_layers}

\subsection{Continuous Relaxation Semantics}

The continuous conjunction (Eq.~\ref{eq:soft_logic1}) approximates AND: when inputs and weights are binary, this reduces to the standard AND: if $\tilde{W}_{ij}^{(\land)} = 1$ (connected), the term becomes $\tilde{n}_j^{(l-1)}$; if $\tilde{W}_{ij}^{(\land)} = 0$ (disconnected), the term becomes $1$, contributing nothing to the product. The output $\tilde{a}_i^{(l)} = 1$ only when all connected inputs equal $1$.

The continuous disjunction (Eq.~\ref{eq:soft_logic2}) leverages De Morgan's law to approximate OR: If any connected input $\tilde{n}_j^{(l-1)} = 1$, the corresponding product term becomes $0$, driving the output toward $1$.

\subsection{Rule Interpretation Procedure}

Post-training, we extract human-readable rules by analyzing learned adjacency matrices $\mathbf{W}^{(\land)}$, $\mathbf{W}^{(\lor)}$ and classifier weights $\mathbf{V}$. For each rule $r_j$, we trace connectivity back through layers to identify the underlying concept combination:

\begin{enumerate}
    \item Identify non-zero entries in $\mathbf{W}^{(\land)}_{j,:}$ or $\mathbf{W}^{(\lor)}_{j,:}$ at the final layer
    \item Recursively trace connections through earlier layers
    \item Construct the logical expression from participating concepts
\end{enumerate}

This yields interpretable rules such as:
\begin{logicbox}
    $r_1 : \texttt{leg\_jump} \land \texttt{arm\_swing} \land \texttt{motion\_upward}$ \\
    $r_2 : \texttt{hand\_grasp} \land \texttt{arm\_raise} \land \neg\,\texttt{leg\_static}$
\end{logicbox}
Combining with classifier weights produces action-level explanations:
\begin{logicbox}
    \textsc{Jump} $\leftarrow 0.82 \cdot r_1 + 0.45 \cdot r_3 - 0.31 \cdot r_7 + \ldots$
\end{logicbox}



\subsection{Logic Layer Hyperparameters}
Ablation studies (Table~\ref{tab:ablations_grid}e) confirm that the default configuration achieves optimal performance, outperforming the small, large and extra large variants.

\begin{table}[h]
\centering
\caption{Logic layer hyperparameters.}
\label{tab:logic_params}
\small
\begin{tabular}{lc}
\toprule
\textbf{Parameter} & \textbf{Value} \\
\midrule
Number of layers $L$ & 2 \\
Conjunction nodes per layer & 128 \\
Disjunction nodes per layer & 128 \\
Skip connections & Yes \\
Negation augmentation & Yes \\
Sparsity weight $\lambda$ & $10^{-6}$ \\
Binarization threshold & 0.5 \\
\bottomrule
\end{tabular}
\end{table}



\section{Training Details}
\label{app:training}

\subsection{Gradient Flow with Grafting and STE}

A key challenge in neurosymbolic learning is maintaining gradient flow through discrete operations. Our framework addresses this through two complementary mechanisms illustrated in Fig.~\ref{fig:model}:

\noindent\textbf{Straight-Through Estimator (STE) for Concept Binarization.}
The binarizer $\bar{c}_i = \mathbb{1}[\hat{c}_i > 0.5]$ produces zero gradients everywhere except at the threshold. The STE bypasses this by copying gradients directly from the binarized output to the soft input:
\begin{equation}
    \text{Forward:} \quad \bar{c}_i = \mathbb{1}[\hat{c}_i > 0.5], \,
    \text{Backward:} \quad \frac{\partial \mathcal{L}}{\partial \hat{c}_i} \coloneqq \frac{\partial \mathcal{L}}{\partial \bar{c}_i}
\end{equation}
This allows the STC-Decoder to receive task-relevant gradients despite the non-differentiable discretization, enabling it to improve concept predictions based on downstream classification performance.

\noindent\textbf{Gradient Grafting for Logic Layers.}
The discrete logic operations ($\land$, $\lor$) and weight binarization similarly block gradients. Gradient grafting maintains two parallel computational streams:
\begin{itemize}
    \item \textbf{Discrete stream (forward pass)}: Binarized concepts $\bar{\mathbf{c}}$ and thresholded weights $\mathbf{W} = \mathbb{1}[\tilde{\mathbf{W}} > 0.5]$ compute rule activations $\mathbf{r}$ via exact Boolean operations. This ensures interpretable, discrete rules during inference.
    \item \textbf{Continuous stream (backward pass)}: Gradients are computed through soft concepts $\hat{\mathbf{c}}$ and continuous weights $\tilde{\mathbf{W}}$ using the differentiable relaxations (Eq.~\ref{eq:soft_logic1} and ~\ref{eq:soft_logic2}). This provides smooth gradient signals for optimization.
\end{itemize}

The ``grafting'' occurs by using the discrete stream's output $\mathbf{r}$ (and thus $\hat{\mathbf{y}}$) for the loss computation, but routing gradients through the continuous stream during backpropagation. Formally:
\begin{equation}
    \frac{\partial \mathcal{L}}{\partial \tilde{W}_{ij}^{(\land)}} = \frac{\partial \mathcal{L}}{\partial \hat{y}} \cdot \frac{\partial \hat{y}}{\partial r} \cdot \frac{\partial \tilde{r}}{\partial \tilde{a}} \cdot \frac{\partial \tilde{a}}{\partial \tilde{W}_{ij}^{(\land)}}
\end{equation}
where $\tilde{r}, \tilde{a}$ denote the continuous relaxations of $r, a$. The key insight is that $\partial \tilde{a} / \partial \tilde{W}_{ij}^{(\land)}$ is well-defined via Eq.~\ref{eq:soft_logic1} and \ref{eq:soft_logic1}, enabling weight updates even though the forward pass used discrete operations.

\subsection{Training Algorithm with Gradient Flow}

Algorithm~\ref{alg:training} details the end-to-end training procedure with explicit gradient flow annotations.

\begin{algorithm}[t]
\caption{End-to-End Training with Gradient Grafting}
\label{alg:training}
\small
\begin{algorithmic}[1]
\REQUIRE Training data $\{(\mathbf{X}_i, a_i)\}$, concept matrix $\mathbf{M}$
\ENSURE Trained model parameters
\STATE Initialize GCN encoder, CLIP (frozen + LoRA), STC-Decoder, Logic Layers
\FOR{epoch $= 1$ to $E$}
    \IF{epoch $\leq 15$}
        \STATE Freeze logic layer weights $\tilde{\mathbf{W}}$ \textcolor{gray}{\textit{// Staggered warmup}}
    \ENDIF
    \FOR{each batch $\{(\mathbf{X}_i, a_i)\}$}
        \STATE \textcolor{gray}{\textit{// Forward pass through encoder and alignment}}
        \STATE $\mathbf{F} \gets \text{GCNEncoder}(\mathbf{X})$
        \STATE $\mathcal{L}_\text{align} \gets \text{ContrastiveLoss}(\text{Pool}(\mathbf{F}), f_\text{text}(\tau_a))$
        \STATE $\mathcal{L}_\text{div} \gets \text{DivergenceLoss}(\mathbf{F})$
        \STATE
        \STATE \textcolor{gray}{\textit{// Concept prediction and supervision}}
        \STATE $\hat{\mathbf{c}} \gets \text{STCDecoder}(\mathbf{F})$ \textcolor{gray}{\textit{// Soft predictions $\in [0,1]^{|\mathcal{C}|}$}}
        \STATE $\mathcal{L}_\text{concept} \gets \text{BCE}(\hat{\mathbf{c}}, \mathbf{M}[a,:])$
        \STATE
        \STATE \textcolor{gray}{\textit{// Discrete forward: binarize concepts (STE backward)}}
        \STATE $\bar{\mathbf{c}} \gets \mathbb{1}[\hat{\mathbf{c}} > 0.5]$ \textcolor{gray}{\textit{// $\nabla_{\hat{\mathbf{c}}} \coloneqq \nabla_{\bar{\mathbf{c}}}$ via STE}}
        \STATE $\mathbf{W} \gets \mathbb{1}[\tilde{\mathbf{W}} > 0.5]$ \textcolor{gray}{\textit{// Discrete weights for forward}}
        \STATE
        \STATE \textcolor{gray}{\textit{// Discrete forward: evaluate logic rules}}
        \STATE $\mathbf{r} \gets \text{LogicLayers}_\text{discrete}(\bar{\mathbf{c}}, \mathbf{W})$ \textcolor{gray}{\textit{// Binary rule activations}}
        \STATE $\hat{\mathbf{y}} \gets \mathbf{V}\mathbf{r} + \mathbf{b}$
        \STATE $\mathcal{L}_\text{task} \gets \text{CrossEntropy}(\hat{\mathbf{y}}, a)$
        \STATE
        \STATE \textcolor{gray}{\textit{// Backward pass: graft gradients through continuous stream}}
        \STATE $\mathcal{L} \gets \mathcal{L}_\text{task} + \alpha \mathcal{L}_\text{concept} + \beta \mathcal{L}_\text{align} + \gamma \mathcal{L}_\text{div} + \lambda \|\tilde{\mathbf{W}}\|_1$
        \STATE Compute $\nabla_{\tilde{\mathbf{W}}} \mathcal{L}$ via continuous relaxations (Eq.~\ref{eq:soft_logic1}and~\ref{eq:soft_logic2})
        \STATE Compute $\nabla_{\hat{\mathbf{c}}} \mathcal{L}$ via STE from $\nabla_{\bar{\mathbf{c}}} \mathcal{L}$
        \STATE Update all parameters via AdamW
    \ENDFOR
\ENDFOR
\RETURN Trained model
\end{algorithmic}
\end{algorithm}

\subsection{Staggered Warmup Strategy}

We employ staggered warmup to prevent spurious rule formation from random concept predictions:
\begin{itemize}
    \item \textbf{Epochs 1-5}: Train GCN encoder, CLIP alignment, STC-Decoder, and part divergence
    \item \textbf{Epochs 1-15}: Logic layers frozen
    \item \textbf{Epochs 16-400}: Full end-to-end training with all components
\end{itemize}

Logic layers use learning rate $10^{-4}$ (10$\times$ higher than other components at $10^{-5}$) to traverse from dense initialization to sparse binary connectivity against $\ell_1$ regularization.

\subsection{Loss Weight Selection}
\label{app:loss_weight_selection}

Our objective combines multiple terms: $\mathcal{L}=\mathcal{L}_\text{task}+\alpha \mathcal{L}_\text{concept}+\beta \mathcal{L}_\text{align}+\gamma \mathcal{L}_\text{div}+\lambda \lVert \tilde{\mathbf{W}}\rVert_1$. We select weights via coordinate-wise sensitivity analysis on NTU RGB+D 120 validation split, sweeping one coefficient at a time. Table~\ref{tab:loss_weight_ablation} shows that moderate alignment ($\beta=0.1$), diversity regularization ($\gamma=1$), and small sparsity penalty ($\lambda=10^{-6}$) yield optimal accuracy with interpretable rules.

\begin{table}[h]
\centering
\small
\caption{Loss weight sensitivity on NTU RGB+D 120 (val.). Sparsity is the number of non-zero rule weights (lower = more interpretable).}
\label{tab:loss_weight_ablation}
\setlength{\tabcolsep}{4pt}
\begin{tabular}{cccccc}
\toprule
$\alpha$ & $\beta$ & $\gamma$ & $\lambda$ & Acc. (\%) & Rules \\
\midrule
1 & 0    & 1 & $10^{-6}$ & 86.4 & 72 \\
1 & 0.05 & 1 & $10^{-6}$ & 87.0 & 61 \\
1 & \textbf{0.1} & \textbf{1} & $\mathbf{10^{-6}}$ & \textbf{87.35} & 44 \\
1 & 0.5  & 1 & $10^{-6}$ & 87.1 & 40 \\
\midrule
1 & 0.1  & 0 & $10^{-6}$ & 86.7 & 58 \\
1 & 0.1  & 2 & $10^{-6}$ & 87.0 & 36 \\
\midrule
1 & 0.1  & 1 & 0         & 87.2 & 95 \\
1 & 0.1  & 1 & $10^{-5}$ & 87.1 & 28 \\
1 & 0.1  & 1 & $10^{-4}$ & 86.5 & 15 \\
\bottomrule
\end{tabular}
\end{table}

\section{Implementation Details}
\label{app:impl_details}

Table~\ref{tab:training_setup} summarizes the overall training configuration, including hyperparameter settings, loss weights, and data augmentation.

\begin{table}[h]
\centering
\small
\caption{Training configuration, loss weights, and data augmentation.}
\label{tab:training_setup}
\begin{tabular}{l c}
\toprule
\textbf{Setting} & \textbf{Value} \\
\midrule
\multicolumn{2}{l}{\textbf{Optimization}} \\
\midrule
Batch size & 64 \\
Epochs & 400 \\
Optimizer & AdamW \\
Weight decay & $10^{-4}$ \\
Gradient clipping & 1.0 \\
Learning rate schedule & Cosine decay \\
\midrule
\multicolumn{2}{l}{\textbf{Learning Rates}} \\
\midrule
GCN / CLIP / STC-Decoder & $10^{-5}$ \\
Logic layers & $10^{-4}$ \\
Logic layers frozen epochs & 15 \\
\midrule
\multicolumn{2}{l}{\textbf{Loss Weights}} \\
\midrule
Task loss ($\mathcal{L}_\text{task}$) & 1.0 \\
Concept supervision ($\alpha$) & 1.0 \\
Skeleton--text alignment ($\beta$) & 0.1 \\
Part divergence ($\gamma$) & 1.0 \\
Rule sparsity ($\lambda \|\tilde{\mathbf{W}}\|_1$) & $10^{-6}$ \\
\midrule
\multicolumn{2}{l}{\textbf{Input \& Augmentation}} \\
\midrule
Input frames & 64 \\
Joints (NTU / NW-UCLA) & 25 / 20 \\
Random rotation & $\pm 15^\circ$ (vertical axis) \\
Spatial flip & 50\% probability \\
Temporal crop & Random 64-frame window \\
Scale jitter & $\pm 10\%$ \\
\bottomrule
\end{tabular}
\end{table}

\section{LLM Re-ranking Details (Optional)}
\label{app:llm_reranking}

The LLM re-ranking module is an \emph{optional} enhancement for scenarios where additional accuracy is prioritized over inference speed. It is not required for the core neurosymbolic framework to function.

\subsection{Motivation}

Recent skeleton-based HAR methods~\cite{ye2025sugar,qu2024llms} leverage LLMs to improve recognition accuracy by incorporating commonsense knowledge about action semantics. However, these methods typically, use generic, sequence-agnostic action descriptions. While requiring two-stage training (pretraining + LLM fine-tuning). Providing limited interpretability (LLM reasoning over opaque features).

Our approach differs by grounding LLM reasoning in \emph{instance-specific} concept activations predicted by our STC-Decoder. This enables the LLM to reason over interpretable evidence (e.g., "\texttt{arm\_swing}: 0.92, \texttt{leg\_jump}: 0.87") rather than abstract feature vectors, maintaining the transparency of our neurosymbolic framework.

\subsection{Prompt Template}

We provide the LLM with predicted concept activations and top-$K$ classifier candidates:

\begin{promptbox}
\texttt{DETECTED MOVEMENT CONCEPTS:}\\
\texttt{~~arm\_swing (0.92), leg\_jump (0.87), motion\_upward (0.85),}\\
\texttt{~~duration\_brief (0.81), feet\_push (0.79), ...}\\[0.5em]
\texttt{CANDIDATE ACTIONS (from classifier):}\\
\texttt{~~1. Jump (score: 2.31)}\\
\texttt{~~2. Hop (score: 1.87)}\\
\texttt{~~3. Skip (score: 1.42)}\\[0.5em]
\texttt{Based on the detected movement concepts, which action is most}\\
\texttt{likely being performed? Only respond with action name.}
\end{promptbox}

\subsection{Key Design Choices}

\noindent\textbf{Instance-specific grounding.} Concept confidences provide concrete evidence from the specific input sequence, enabling distinction of execution variants (e.g., arms-first vs. legs-first jump).

\noindent\textbf{Candidate narrowing.} Providing top-$K$ candidates ($K=5$) constrains the LLM output space, improving reliability and reducing hallucination.

\noindent\textbf{Preserved interpretability.} The LLM selects among candidate actions based on human-readable concept activations, maintaining the auditability of our framework.

\subsection{Fine-tuning Configuration}

After training the core framework, we freeze all modules and fine-tune LLaMA-3.2 7B with LoRA~\cite{hu2022lora}.

\begin{table}[h]
\centering
\caption{LLM re-ranking hyperparameters.}
\label{tab:llm_params}
\small
\begin{tabular}{lc}
\toprule
\textbf{Parameter} & \textbf{Value} \\
\midrule
Base model & LLaMA-3.2 7B \\
LoRA rank & 16 \\
LoRA $\alpha$ & 32 \\
Learning rate & $2 \times 10^{-5}$ \\
Batch size & 4 \\
Epochs & 2 \\
Top-$K$ candidates & 5 \\
\bottomrule
\end{tabular}
\end{table}

\section{Concept Intervention Analysis}
\label{app:intervention}

\subsection{Motivation}

A key advantage of neurosymbolic models is their amenability to human intervention: if a concept is mispredicted, a user can correct it and observe the downstream effect on the action prediction. This property enables debugging, trust calibration, and human-in-the-loop refinement, capabilities absent in black-box models where internal representations are opaque and non-editable.

\subsection{Experimental Protocol}

We simulate concept intervention by replacing predicted concept activations $\hat{c}_i$ with ground-truth values $c^*_i$ derived from the action-concept matrix $\mathbf{M}$. Specifically, we:
\begin{enumerate}
    \item Identify samples where the model prediction is incorrect
    \item Rank predicted concepts by their deviation from ground truth: $|\hat{c}_i - c^*_i|$
    \item Progressively correct concepts starting from the most erroneous
    \item Re-evaluate the action prediction after each correction
\end{enumerate}

\subsection{Results and Interpretation}

Table~\ref{tab:ablations_grid}(f) reports results on NTU-120 X-Set. Correcting a single mispredicted concept improves accuracy by 1.88\%, with gains scaling monotonically to 5.67\% when three concepts are corrected.

This consistent improvement demonstrates two important properties: (i) The logic layers have learned meaningful rules, correcting upstream concept errors propagates meaningfully through the reasoning chain to improve predictions, and, (ii) Prediction failures are predominantly attributable to concept misprediction rather than flawed logical rules. This suggests that improving the STC-Decoder's concept prediction accuracy would yield further gains.

\textbf{Limitations:} The intervention analysis assumes access to ground-truth concept labels, which may not be available in deployment. However, the key insight, that concept corrections improve predictions, remains valid for human-provided corrections based on visual inspection of the skeleton sequence. Future work could explore active learning strategies that query humans for concept corrections on low-confidence predictions.

\begin{figure}[h]
    \centering
    \includegraphics[width=0.6\linewidth]{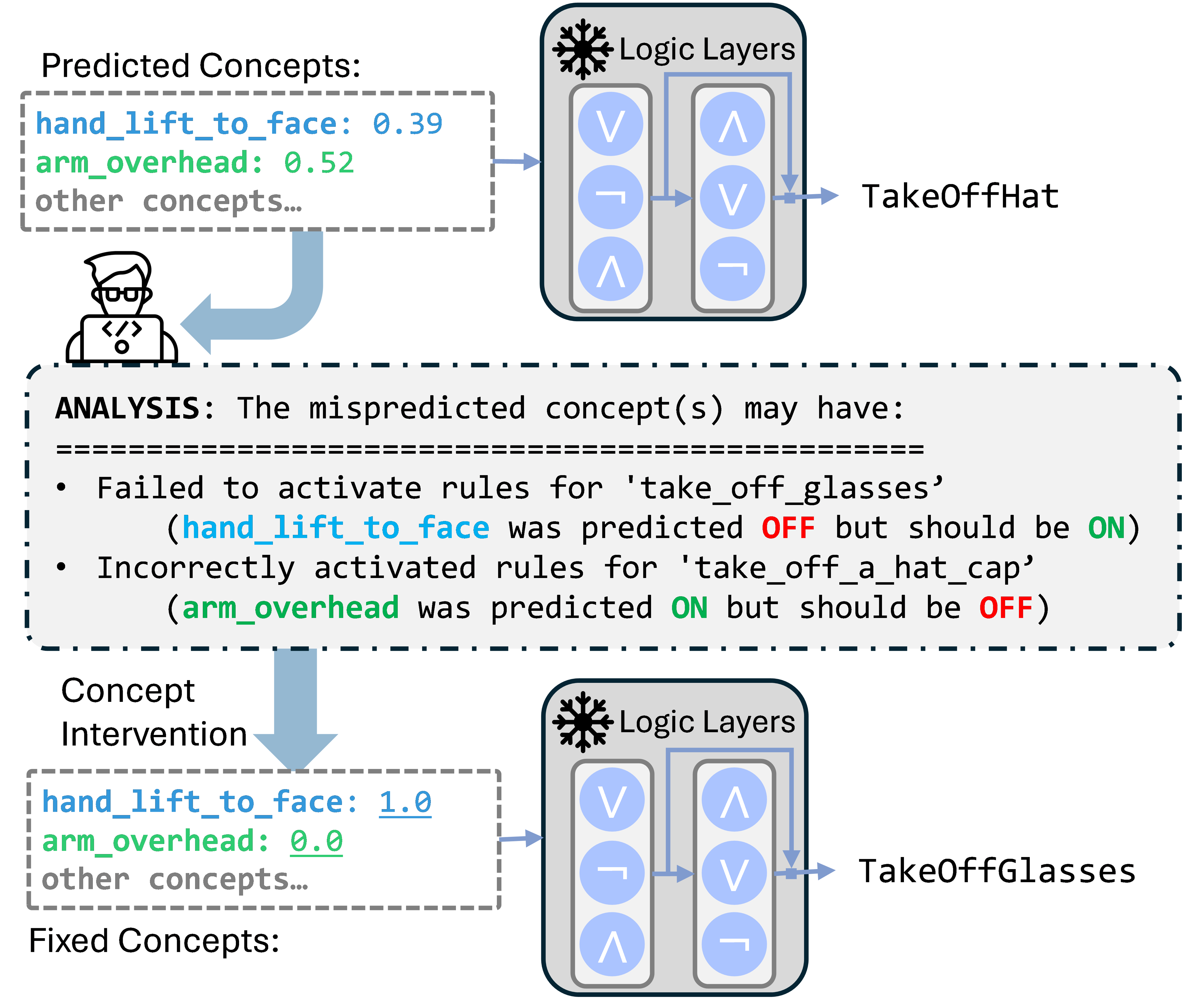}
    \caption{Overview of the concept intervention pipeline, illustrating how correcting mispredicted concepts propagates through symbolic reasoning to fix action predictions.}
    \label{fig:concep_interv}
\end{figure}

\section{Additional Visualizations}
We provide additional qualitative visualizations to further illustrate how our model grounds skeleton sequences into interpretable spatio-temporal concepts and composes them into logical rules.
\begin{figure}[!h]
    \centering
    \includegraphics[width=1.0\linewidth]{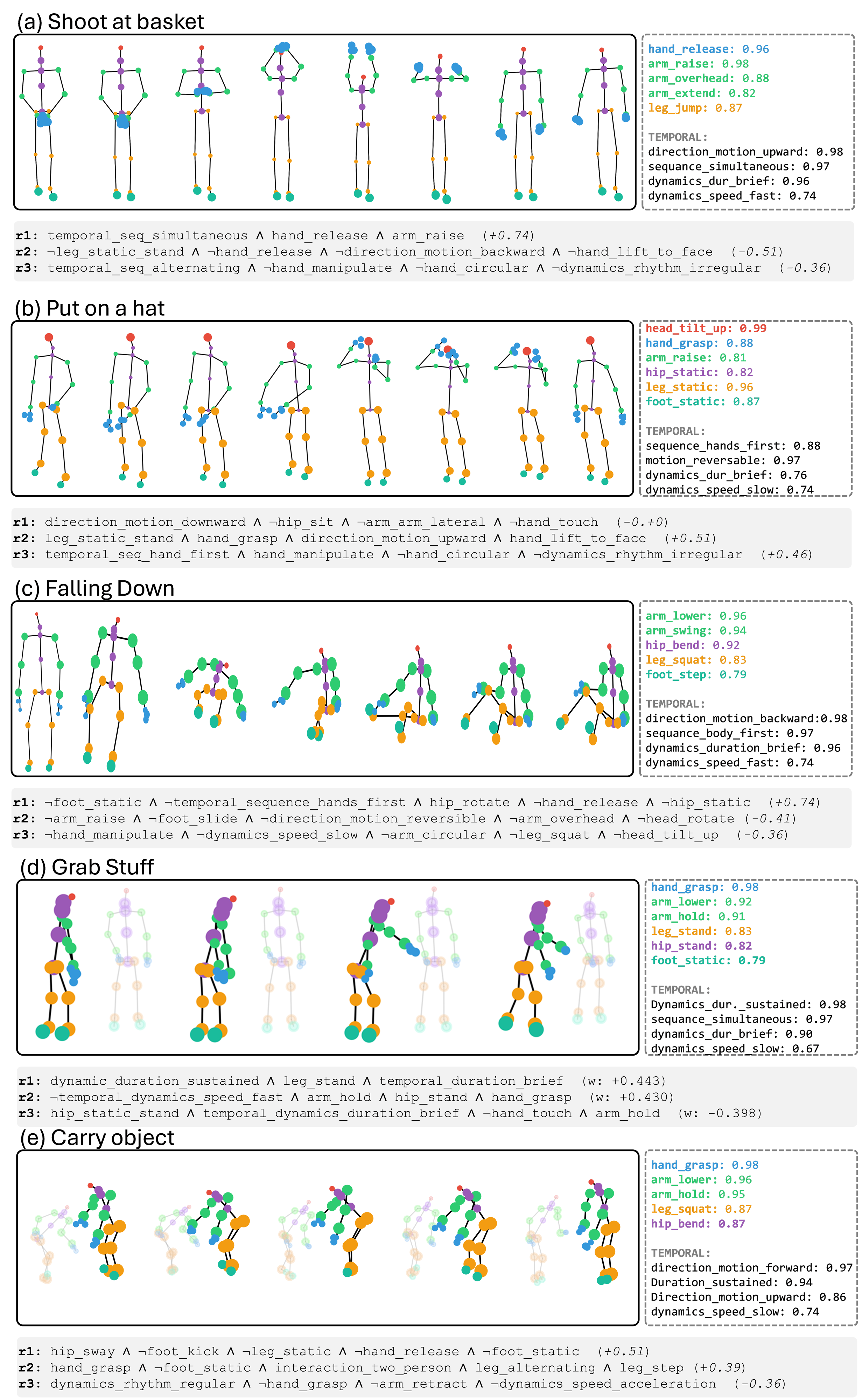}
    \caption{\textbf{Visualization of our concept-grounded neurosymbolic reasoning.} Each panel shows temporally sampled skeletons with concept activations, top spatial–temporal concepts (dotted box), and top-3 learned logical rules with weights (gray box).}
    \label{fig:skeleton_app}
\end{figure}

\section{Concept Activation Analysis}
\label{app:concept_analysis}

To validate that REASON learns semantically meaningful concepts, we analyze activation patterns across the NTU RGB+D 120 action classes grouped by semantic category (Table~\ref{tab:action_groups}).

\subsection{Concept Sparsity and Discriminability}

Figure~\ref{fig:heatmap} presents a grouped action-concept activation heatmap with coefficient of variation (CoV) and mean activation statistics. Several key findings emerge:

\begin{figure*}[h]
    \centering
    \includegraphics[width=0.95\linewidth]{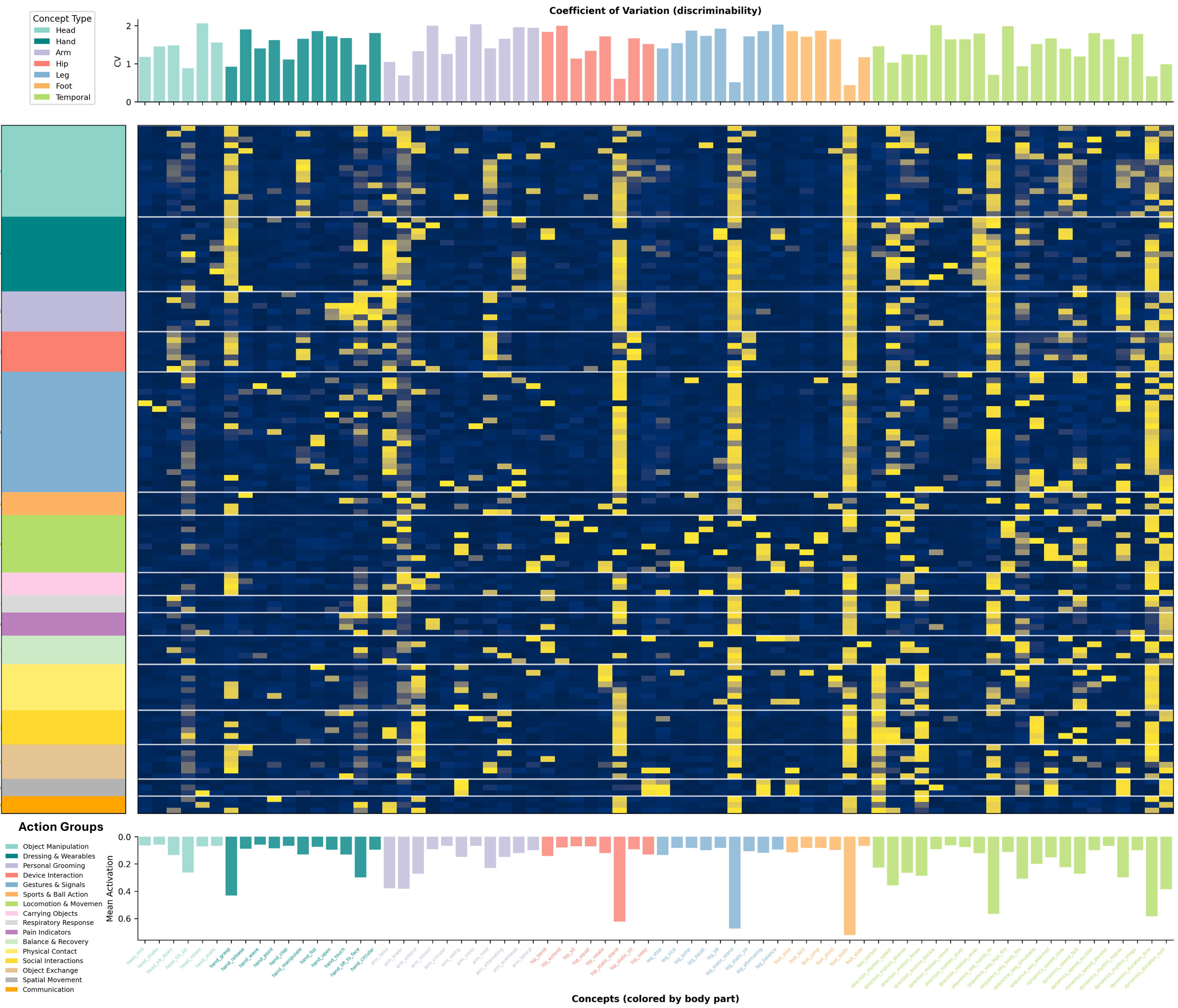}
    \caption{\textbf{Grouped action-concept activation heatmap} on NTU RGB+D 120 (X-Set validation). Top: Coefficient of variation (CoV) per concept: higher CoV indicates greater discriminative power. Bottom: Mean activation frequency. Column-wise normalization emphasizes relative concept importance within each action group. Temporal concepts (right) exhibit slightly higher activation and CoV than spatial concepts (left).}
    \label{fig:heatmap}
\end{figure*}

\noindent\textbf{Sparse activations.} On average, only 8 concepts activate per action (out of 67), indicating that the framework learns compact, interpretable representations rather than dense, redundant encodings. This sparsity aligns with human intuition, actions are typically characterized by a few salient motion primitives.

\noindent\textbf{Discriminative vs.\ common concepts.} The CoV analysis (top of heatmap) reveals which concepts discriminate between action classes versus those that activate broadly. Static concepts (\texttt{foot\_static}, \texttt{leg\_static}, \texttt{hip\_static}) exhibit low CoV despite high mean activation, indicating they are frequently present but contribute minimally to discrimination. In contrast, concepts with high CoV are selectively activated for specific action groups, providing stronger discriminative signal.

\noindent\textbf{Temporal concept importance.} Temporal concepts (right side of heatmap) show both higher mean activation and higher CoV compared to spatial concepts (left side). This empirically validates our design choice to explicitly model temporal dynamics, temporal concepts are not only more frequently relevant but also more discriminative, consistent with our ablation finding that temporal concepts contribute +7.8\% accuracy improvement.

\subsection{Action Group Profiles}

\begin{figure*}[h]
    \centering
    \includegraphics[width=0.8\linewidth]{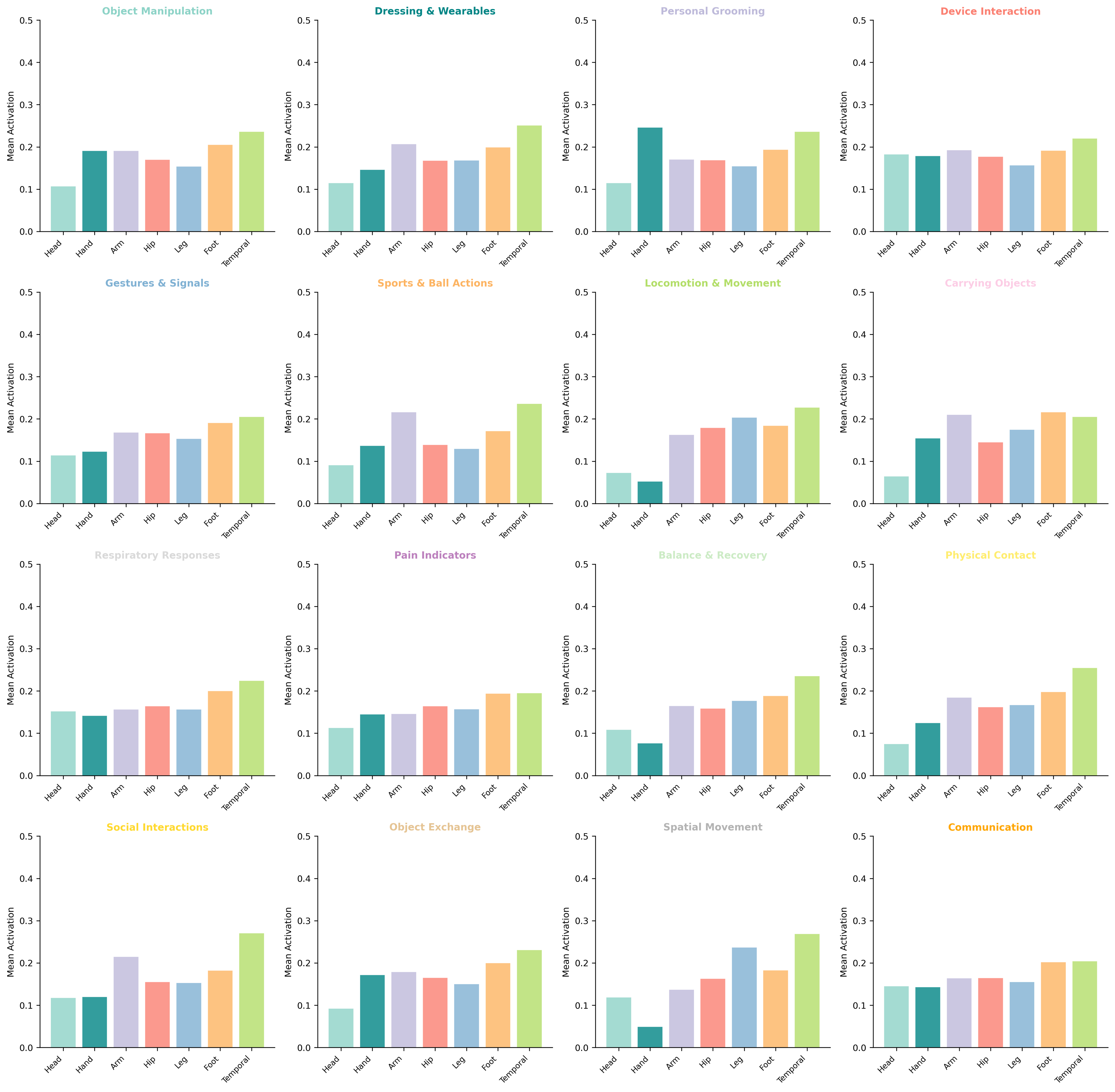}
    \caption{\textbf{Body-part concept activation profiles by action group.} Each bar shows the mean activation of concepts associated with a specific body part or temporal category. Action groups exhibit intuitive profiles: grooming actions activate hand concepts, locomotion activates leg/foot concepts, and sports actions show elevated temporal dynamics.}
    \label{fig:act_profiles}
\end{figure*}

Figure~\ref{fig:act_profiles} visualizes body-part concept activation profiles for each semantic action group, revealing intuitive correspondences:

\begin{itemize}
    \item \textbf{Personal Grooming} (brushing teeth, applying cream): Highest \texttt{hand} concept activations, reflecting fine manipulation near the face.
    \item \textbf{Sports \& Ball Actions}: Elevated \texttt{hand} and \texttt{temporal} activations, capturing both object interaction and dynamic motion patterns.
    \item \textbf{Locomotion \& Movement}: Dominant \texttt{leg}, \texttt{foot}, and \texttt{temporal} activations, encoding lower-body motion and movement dynamics.
    \item \textbf{Spatial Movement} (walking towards/apart): Strong \texttt{leg}, \texttt{foot}, and \texttt{temporal} profiles, distinguishing approach from departure via directional temporal concepts.
\end{itemize}

These profiles demonstrate that REASON's learned concepts align with human understanding of action semantics, body-part concepts activate according to the anatomical regions involved, while temporal concepts capture the characteristic dynamics of each action category.

\subsection{Discriminative Concepts for Similar Actions}

\begin{figure*}[h]
    \centering
    \includegraphics[width=0.9\linewidth]{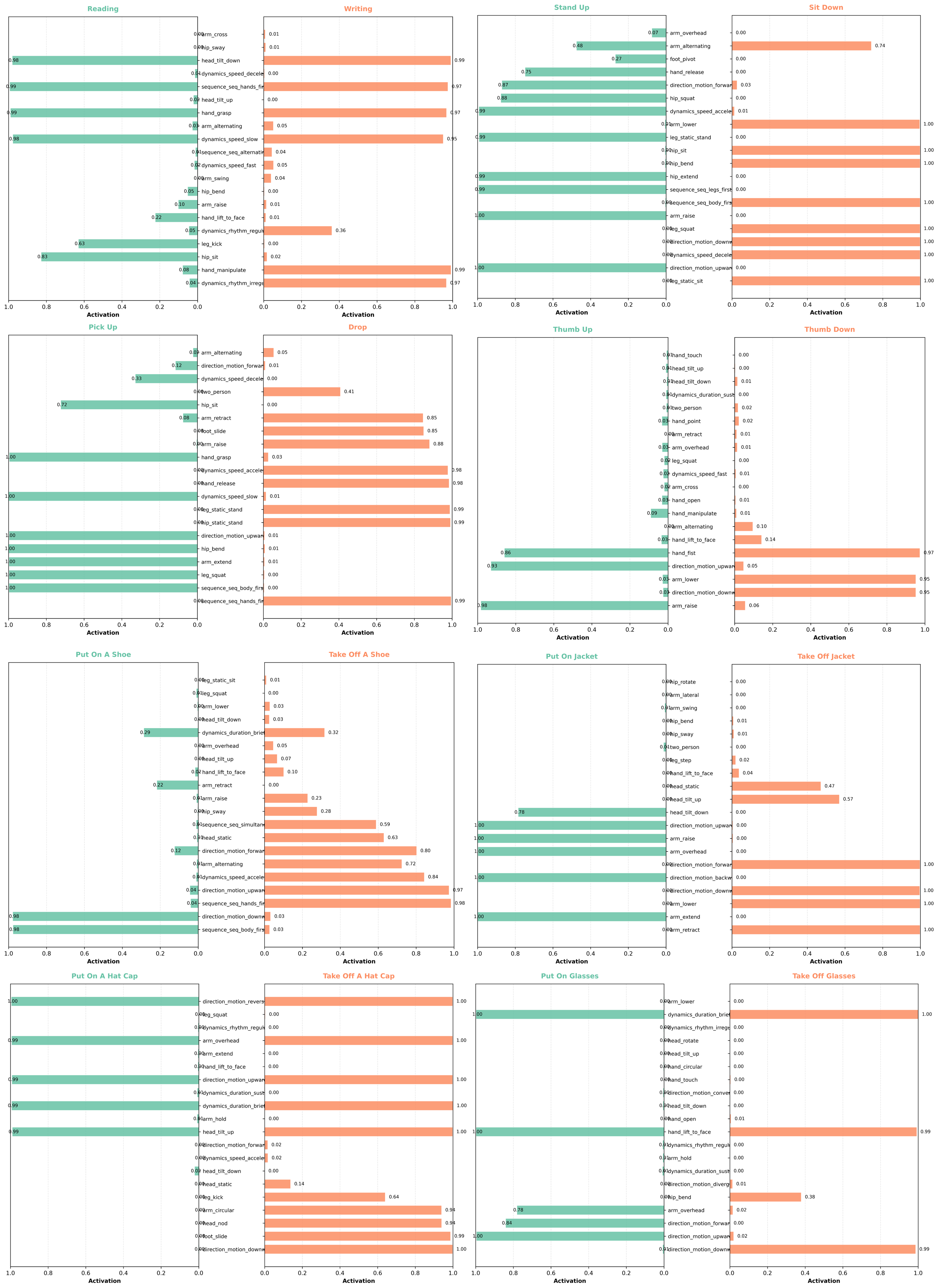}
    \caption{\textbf{Concept activation comparison for semantically similar action pairs.} Top-20 most discriminative concepts are highlighted. Temporal concepts (e.g., \texttt{motion\_forward} vs.\ \texttt{motion\_backward}) often distinguish action pairs that share similar spatial configurations.}
    \label{fig:butter_1}
\end{figure*}

Figure~\ref{fig:butter_1} compares concept activations for eight pairs of semantically similar actions, highlighting the top-20 most discriminative concepts. For challenging pairs like \textsc{Put on Shoe}/\textsc{Take off Shoe}, the distinguishing concepts are primarily temporal (\texttt{motion\_forward} vs.\ \texttt{motion\_backward}), while pairs like \textsc{Reading}/\textsc{Writing} differ in hand manipulation concepts (\texttt{hand\_hold} vs.\ \texttt{hand\_write}). This analysis confirms that the concept vocabulary captures the subtle motion differences needed to distinguish semantically similar actions—the core challenge our framework addresses.

The concept activation analysis provides empirical evidence that REASON learns sparse, semantically grounded representations. Concepts align with human intuition about action semantics, temporal concepts provide critical discriminative power, and the learned vocabulary captures subtle differences between similar actions—validating the interpretability claims of our neurosymbolic framework.

\begin{table*}[h]
\centering
\small
\caption{\textbf{Semantic grouping of NTU RGB+D action classes} used for concept activation analysis. Actions are grouped by functional similarity to reveal body-part and temporal concept patterns.}
\label{tab:action_groups}
\begin{tabular}{p{3.2cm} p{12.3cm}}
\toprule
\textbf{Action Group} & \textbf{Action Classes} \\
\midrule
Object Manipulation &
drink\_water, eat\_meal, drop, pick\_up, throw, tear\_up\_paper, staple\_book, counting\_money, cutting\_nails, cutting\_paper, open\_bottle, open\_a\_box, toss\_a\_coin, fold\_paper, ball\_up\_paper, play\_magic\_cube \\
Dressing \& Wearables &
put\_on\_jacket, take\_off\_jacket, put\_on\_a\_shoe, take\_off\_a\_shoe, put\_on\_glasses, take\_off\_glasses, put\_on\_a\_hat\_cap, take\_off\_a\_hat\_cap, put\_on\_headphone, take\_off\_headphone, put\_on\_bag, take\_off\_bag, throw\_up\_cap\_hat \\
Personal Grooming &
brush\_teeth, brush\_hair, wipe\_face, apply\_cream\_on\_face, apply\_cream\_on\_hand, flick\_hair, sniff\_smell \\
Device Interaction &
reading, writing, phone\_call, play\_with\_phone\_tablet, type\_on\_a\_keyboard, taking\_a\_selfie, check\_time\_(from\_watch) \\
Gestures \& Signals &
clapping, cheer\_up, hand\_waving, point\_to\_something, rub\_two\_hands, nod\_head\_bow, shake\_head, salute, put\_palms\_together, cross\_hands\_in\_front, hush, thumb\_up, thumb\_down, make\_OK\_sign, make\_victory\_sign, snap\_fingers, shake\_fist, capitulate, cross\_arms, arm\_circles, arm\_swings \\
Sports \& Ball Actions &
shoot\_at\_basket, bounce\_ball, tennis\_bat\_swing, juggle\_table\_tennis\_ball \\
Locomotion \& Movement &
sit\_down, stand\_up, squat\_down, hopping, jump\_up, run\_on\_the\_spot, butt\_kicks, cross\_toe\_touch, side\_kick, kicking\_something \\
Carrying Objects &
reach\_into\_pocket, put\_object\_into\_bag, take\_object\_out\_of\_bag, move\_heavy\_objects \\
Respiratory Responses &
sneeze\_cough, blow\_nose, yawn \\
Pain Indicators &
headache, chest\_pain, back\_pain, neck\_pain \\
Balance \& Recovery &
staggering, falling\_down, nausea\_vomiting, fan\_self, stretch\_oneself \\
Physical Contact &
punch\_slap, kicking, pushing, hit\_with\_object, wield\_knife, shoot\_with\_gun, step\_on\_foot, knock\_over \\
Social Interactions &
pat\_on\_back, hugging, shaking\_hands, high-five, cheers\_and\_drink, rock-paper-scissors \\
Object Exchange &
giving\_object, exchange\_things, grab\_stuff, support\_somebody, carry\_object, touch\_pocket \\
Spatial Movement &
walking\_towards, walking\_apart, follow \\
Communication &
point\_finger, whisper, take\_a\_photo \\
\bottomrule
\end{tabular}
\end{table*}

\end{document}